\documentclass[final,5p,times,twocolumn]{elsarticle}

\usepackage{diagbox}
\usepackage{multirow}
\usepackage[table]{xcolor}
\usepackage{times}
\usepackage{fancyhdr, graphicx, amsmath, amssymb}

\usepackage{algorithmic}
\usepackage{algorithm}

\makeatletter
\patchcmd{\algorithm}{\fs@ruled}{\fs@plain}{}{} 
\patchcmd{\endalgorithm}{\endcenter}{\endcenter\vspace{-1.5\baselineskip}}{}{} 
\makeatother

\usepackage{nomencl}
\makenomenclature
\usepackage{tcolorbox}
\usepackage{tikz}
\tcbuselibrary{breakable}
\newtcolorbox{nomenclBox}{%
    breakable,          
    arc=0pt,           
    boxrule=1.0pt,   
    colback=white,      
    colframe=black,    
    left=10pt,          
    right=10pt,  
}

\usepackage{subcaption} 
\usepackage{arydshln}

\definecolor{tablegray}{gray}{0.8}

\usepackage{booktabs}
\usepackage{makecell}
\usepackage{array}   
\usepackage{adjustbox} 

\usepackage{siunitx}    
\usepackage{amsmath}    
\usepackage{euscript}    

\usepackage{array}
\usepackage{multirow}
\usepackage{tabularx}
\usepackage{hyperref}
\usepackage{caption}
\journal{Applied Energy}

\begin{document}
\include{pythonlisting}
\begin{frontmatter}



%\title{State-Action Representation Learning Enhanced Deep Reinforcement Learning for Distributed Energy System Scheduling and Optimization\tnoteref{t1}}

\title{Representation Learning Enhanced Deep Reinforcement Learning for Optimal Operation of Hydrogen-based  Multi-Energy Systems\tnoteref{t1}}

% \title{Reinforcement Learning vs Model Predictive Control: A Comparative Study in Hydrogen-Integrated Energy Systems Optimization}
% \title{Comparative Study of RL and MPC for Operation Optimization in Hydrogen-Integrated Energy Systems}
\tnotetext[t1]{This  work  is  supported by the National Natural Science Foundation of China  (62403373, 72595834, 62192752, 62125304) and 111 International Collaboration Program (B25027).}

\author[1]{Zhenyu Pu}
\ead{zhenyupu@stu.xjtu.edu.cn}

\author[1]{Yu Yang\corref{cor1}}
\ead{yangyu21@xjtu.edu.cn}
\cortext[cor1]{Yu Yang is corresponding author.}

\author[1]{Lun Yang\corref{cor1}}
\ead{yanglun@xjtu.edu.cn}

\author[2]{Qing-Shan Jia}
\ead{jiaqs@tsinghua.edu.cn}

\author[1,2]{Xiaohong Guan}
\ead{xhguan@tsinghua.edu.cn}

\author[3]{Costas~J.~Spanos}
\ead{spanos@berkeley.edu}

%\fntext[fn2]{Another author footnote}

\affiliation[1]{organization={School of Automation Science and Engineering, Xi’an Jiaotong University},
        addressline={Shaanxi},
        city={Xi'an},
        postcode={710049},
        country={China}
    }

\affiliation[2]{organization={Center for Intelligent and Networked Systems,
		Department of Automation, BNRist, Tsinghua University},
	    addressline={Beijing},
        postcode={100084},
        country={China}
}

\affiliation[3]{organization={Department of Electrical Engineering and 	Computer Sciences, University of California},
	addressline={Berkeley},
    city={CA},
	postcode={94720},
	country={USA}
}

\begin{abstract}
Hydrogen-based multi-energy systems (HMES) have emerged as a promising low-carbon and energy-efficient  solution, as  it can enable the coordinated operation of electricity, heating and cooling supply and demand  to enhance  operational flexibility, improve overall energy efficiency, and increase the share of renewable integration. However, the optimal operation of HMES remains challenging due to the nonlinear and multi-physics coupled dynamics of hydrogen energy storage systems (HESS) (consisting of electrolyters,  fuel cells and hydrogen tanks) as well as  the presence of multiple uncertainties from supply and demand. To address these challenges, this paper develops a comprehensive  operational model for HMES that fully captures the nonlinear dynamics and multi-physics process of HESS. Moreover, we propose an enhanced deep reinforcement learning (DRL) framework by integrating the emerging  representation learning techniques, enabling substantially accelerated and improved policy optimization  for  spatially and temporally coupled complex networked systems, which is not provided by conventional DRL. Experimental studies based on real-world datasets show  that the comprehensive model is crucial  to ensure the safe and reliable of HESS. In addition, the proposed SR-DRL approaches demonstrate superior  convergence rate and  performance over conventional DRL counterparts in terms of reducing the operation cost of HMES and handling the system operating constraints. Finally, we  provide some insights into the role of representation learning in DRL, speculating that it can  reorganize the original  state space into a well-structured and cluster-aware geometric representation, thereby smoothing and facilitating the learning process of DRL.
\end{abstract}


% \begin{highlights}
% \item \textbf{A comprehensive operational model for hydrogen-based multi-energy system (HMES):}  Established  a comprehensive model for the optimal operation of HMES that  fully incorporates the nonlinear operating characteristics and  multi-physics coupled process of hydrogen-based energy system (HESS), which is crucial to  ensure  reliable and efficient system operation. 
% \item \textbf{An enhanced Deep Reinforcement Learning (DRL) approach for complex networked dynamic systems:} Proposed a novel state-action representation learning enhanced deep reinforcement learning (SR-DRL) framework to  address the challenges of learning-based control for complex systems,  significantly improving  policy convergence and performance  over conventional DRL. 
% \item  \textbf{Some illustrative insights of representation learning for enhancing deep reinforcement learning (DRL)}: Provide some insights into the role of the state–action representation learning (SR)  for DRL and conclude that it can reorganize the original state-space of HMES into a well-structured and cluster-aware presentation, thereby facilitating actor–critic learning in DRL methods.
% \end{highlights}

\begin{keyword}
Distributed energy systems (DES) \sep Representation learning \sep Deep reinforcement learning (DRL) \sep Accelerated Learning \sep Hydrogen energy storage \sep decision-making for complex networked system
%%% MSC codes here, in the form: \MSC code \sep code
%%% or \MSC[2008] code \sep code (2000 is the default)
\end{keyword}

\end{frontmatter}


\section{Introduction}
\label{sec:introduction}
Driven by the severe pressure of greenhouse gas effects and fossil fuel depletion, global energy sector is undergoing a profound transition toward low-carbon and sustainability. The transition  had led to the proliferation of  distributed energy systems  that integrate on-site generation, energy conversion, storage, and management units to enable local energy supply with high share of renewables and enhanced  energy efficiency.
Among various emerging distributed energy systems, hydrogen-based multi-energy systems have attracted  growing attention due to many unique advantages of hydrogen as an energy carrier \cite{RL2026CriticalReview}.
\emph{First}, hydrogen has a wide range of sources, including renewable-powered  water electrolysis,  industrial by-products and fossil-fuel based reforming processes, providing highly flexible and abundant sourcing.  
\emph{Second}, hydrogen enables large-scale,  long-duration storage and transport with relatively low self-discharge rate,  making it particularly suitable for seasonal, cross-spatial and cross-temporal energy balancing, which is not provided by other types of energy carriers.  \emph{Third}, hydrogen exhibits prominent  capability for multi-energy and cross-sector integration, as it enables flexible conversion among diverse energy carriers and coordinated operation across energy sectors. As an intermediate energy carrier, hydrogen can be converted into electricity, heat, and cooling.  Moreover, it supports a wide range of applications, including transportation for fuel cell vehicles, buildings for heating and cooking, and power systems for generation and storage.

Hydrogen-based multi-energy systems  typically incorporate  hydrogen energy storage systems (HESS), which consist of electrolyters (ELs),  fuel cells (FCs) and hydrogen tanks (HTs),  enabling flexible energy conversion and storage. Specifically,  ELs produce hydrogen through water electrolysis, which is subsequently stored in hydrogen tanks. When required, the stored hydrogen is converted by fuel cells into electricity and heat. HESS has experienced rapid technological progress in recent decade~\cite{RL2026CriticalReview}. Several types of electrolyzers have been developed, including  Alkaline Electrolyzers (AEL), Proton Exchange Membrane (PEM) Electrolyzers, Solid Oxide Electrolyzers (SOEC) and Anion Exchange Membrane (AEM) Electrolyzer. 
Among them, the emerging PEM electrolyzers are particularly suitable for  renewable-powered  hydrogen production as they have quite fast dynamic response, relatively high energy efficiency and low operating temperatures.
For hydrogen storage, high-pressure gas hydrogen storage (typically at 350--700 bar) and liquid hydrogen storage (requiring cryogenic temperatures below -253°C) are commonly adopted. 
For fuel cells, Proton Exchange Membrane Fuel Cells (PEMFCs) and Solid Oxide Fuel Cells (SOFCs) are  two dominant types, in which the PEMFCs are  well suited for distributed energy systems due to their relatively  low operating temperatures (approximately 70--80 °C) and fast start-up characteristics.

Hydrogen-based multi-energy systems (HMES) have attracted extensive attention from the research community. Existing studies are spreading system architecture designing,  capacity planing and optimal operation.  
In terms of system design, Wang et al.~\cite{multi-obj-hydrogen-collective} proposed an HMES that integrates natural gas and solar energy supply with various  generation, conversion, and storage technologies to achieve coordinated multi-energy provision for building communities, with its economic, environmental, and reliability advantages over conventional  independent operation  demonstrated by case studies.
 Zhang et al.~\cite{flamm2021electrolyzer} investigated  a  hydrogen-based multi-carrier energy system  that incorporates  renewable generation,
electricity and hydrogen markets for a jointly provision of power, cooling, and heating.  
Liu et al.~\cite{LiuJH} studied  comprehensive  capacity planning for similar HMES and compared its economic benefits for different operating scenarios.
The above  studies generally  focus on system design and planning considering  the economic and environmental benefits of HMES.

The optimal operation of HMES requires to formulate and solve  comprehensive optimization problems that involve multi-energy couplings, temporal dependencies, diverse energy device operating characteristics, and multiple uncertainties regarding  supply and demands, which have attracted widespread attention. We have investigated and summarized existing studies in Table~\ref{tab:literature},  highlighting their application scenarios, modeling approaches, uncertainty management, solution methods, as well as  advantages and limitations.  
For application scenarios, prior works have investigated HMES for communities, microgrids, buildings and electric vehicle (EV) charging stations.  Different types of models have been employed to characterize the operating behaviors of HESS. Linear models typically assume constant energy conversion efficiency across the operating ranges of ELs anf FCs. In contrast, nonlinear models explicitly account for the variation in energy conversion efficiency of ELs and FCs. Commonly-used solution methods  include  stochastic programming (SP), model predictive control (MPC) and deep reinforcement learning (DRL).  They  adopt different approaches to address uncertainties. SP methods commonly rely on scenario-based representaion to capture the daily uncertainties of renewable generation and energy demands, while MPC approaches generally depend on forecast or scenario-based chance constraints. In contrast, DRL methods  learn optimal control policies under uncertainties from experience.

Overall, some  existing studies (i.e.,  \cite{dong2022optimal, PeakShaving, dong2023refined,  qi2025long}) have paid attention to the intrinsic nonlinear operating dynamics of HESS, which are essential for  reliable and efficient system operation.  However, these works primarily  rely  on  SP  and  stochastic MPC and suffer from high computational complexity due to the need to solve constrained multi-stage nonlinear stochastic optimization problems. Typically, Dong et al.~\cite{dong2022optimal} studied the optimal operation of an HMES using scenario-based SP. The nonlinear behaviors of FC and absorption chiller (AC) were considered and  handled by piecewise linearization. Nevertheless, this approach remains computational challenging  due to the large number of  mixed-integer constraints and representative scenarios for characterizing uncertainties.

Considering the system complexity and multiple uncertainties, deep reinforcement learning (DRL) has emerged as a promising solution for the optimal operation of HMES.  Unlike optimization-based approaches, DRL learns optimal control strategies from the operating experience of systems, making it advantageous to accommodate  the nonlinear and complex operating behaviors of HESS. In recent years, many DRL methods have been exploited for  HMES. Typically, Yu et al.~\cite{yu2022joint} studied the optimal operation of a building HMES  based on MAADDPG. Zhu et al.~\cite{zhang2023multi} studied the  joint operation of interconnected hydrogen-based multi-energy microgrids  based on multi-agent DRL.  
 Dolatabali et al.~\cite{dolatabadi2022novel} has explored the combination of DDPG  with CNN and LSTM to enable  the optimal operation of an  HMES.  
 The CNN and LSTM were employed to process the high-dimensional input features associated with cloud-related weather conditions.  These works have demonstrated the potential and advantage of  DRL methods the optimal control of complex dynamics systems under uncertainties.  However, many of the widely-used conventional DRL methods are not effective enough to accommodate the system complexity and strong spatial-temporal couplings caused by multi-energy flows and coordinated operation of diverse energy generation, conversion and storage devices. They often suffer from slow convergence and lead to deficient control policies.  
Specifically, conventional DRL-based control policies  may exhibit a  performance gap of 40\%+ relative to the theoretical optimum \cite{yu2022joint}.  Besides, most existing works relied on simplified linear models for HESS, without considering the intrinsic nonlinear operating characteristics and  multi-physics coupled process of ELs and FCs due to computational limitations. 	Consequently, the resulting control policies  deviate substantially from practical operating conditions and even violate device operating limits.  As reported in a recent experimental study \cite{flamm2021electrolyzer}, both ELs and FCs exhibit  
 pronounced nonlinear and multi-physics coupled behaviors. Specifically, the energy conversion efficiency and operational safety of HESS are jointly influenced by the coupled thermal, electrochemical, and pressure dynamics of ELs, FCs, and hydrogen tanks.   Notably, increasing recent works have emphasized the importance to account for the nonlinear operating characteristics and  multi-physics coupled process  to ensure reliable and efficient operation of HESS. For example,  \cite{dong2023refined}  proposed an advanced operating model for an Electric-Hydrogen-Thermal-Gas Integrated Energy System  and demonstrated a 3\% annual cost reduction by accounting for the nonlinear operating characteristics of HESS.  Marta et al.~\cite{PeakShaving} enabled the coordinated operation of a practical HESS and an electric vehicle charging station (EVCS) for peak demand shaving with  the multi-physics coupled dynamics of HESS considered.

From the literature, we note that for the optimal operation of HMES, two critical problems remain to be addressed: (1) build a comprehensive model that considers  the nonlinear   and multi-physics coupled operating behaviors of HESS to ensure reliable and efficient operation of  HMES. (2) handle the computational complexity  caused by the system complexity and the multiple uncertainties.

\renewcommand{\arraystretch}{1.1} 
\begin{table*}[htbp] 
	\centering
	\caption{Existing works of optimal operation of hydrogen-based multi-energy system (HMES)}
	\setlength{\tabcolsep}{1.8pt}
	\label{tab:literature}
	\begin{tabular}{lllcllccc} 	
		\toprule
		\makecell[c]{Scenarios}              
		& Models   
		& \makecell[c]{Uncertainties\\Handing}            
		& \makecell[c]{Solution \\ methods}                         
		%		& \makecell[c]{Method\\Scalablity}  
		&Advantages
		&Limitations
		& Year 
		& Paper  \\
		\hline    
		Community  &   \makecell[c]{Nonlinear\\ model}  & \makecell[l]{Representative\\ scenarios}  & SP  & \makecell[l]{Capture nonlinear \\ operating characteristics \\ of HESS}    & \makecell[l]{High computational\\  burden due to nonlinear\\  models and \\ scenario-based \\ uncertainties}       & 2022       & \cite{dong2022optimal}  \\
		\hline
				\makecell[l]{EV\\charging\\station}     
		& \makecell[c]{Linear\\model}           
		& Scenario trees      & SP   &  \makecell[l]{Mitigate forecast \\ deviations  by \\ two-stage stochastic \\ scheduling } &     \makecell[l]{High computational  \\ burden due to  \\ uncertainties;  \\ Strong dependence \\ on forecast accuracy.}    & 2021     & \cite{mei2021stochastic}  \\[4pt]
		\hline 
		Microgrid
		& \makecell[c]{Nonlinear\\model}           
		& \makecell[l]{Representative \\ scenarios}
		&  SP
		& \makecell[l]{Accounting for  \\ nonlinear behaviors\\
			of HEMS}                 
		& \makecell[l]{High computational \\ complexity due to \\ model complexity  \\and  uncertainties}      
		& 2025         & \cite{qi2025long} \\
				\hline
		Microgrid    & \makecell[c]{Nonlinear\\model}            & Deterministic       & MILP &  \makecell[l]{Consider the nonlinear \\ operating  characteristics \\  of HESS.}      & \makecell[l]{Computational complexity\\  due to model complexity;\\ Do not account for \\ multiple uncertainties}        & 2023    & \cite{dong2023refined}  \\[4pt]
		\hline 
		Microgrid     & \makecell[c]{Linear\\ model}  &  \makecell[l]{Multi-time-scale \\ forecast of load \\ and supply}  & MPC                & \makecell[l]{Consider multi-time-scale \\ characteristics of multiple\\ energy demands}     & \makecell[l]{Ignores nonlinear \\ characteristics and \\ safety constraints \\ of HESS }       & 2022   & \cite{fang2022multiple} \\
		\hline
		\makecell[l]{EV \\Charging \\ Station}         & \makecell[l]{Nonlinear\\ model}       & \makecell[l]{Probabilistic \\ uncertainties \\ with chance\\ constraints}  & \makecell[c]{Stochastic \\ MPC} & \makecell[l]{Account for multi-physics \\ process of HESS \\ and  safe constraints}                   & \makecell[l]{High computational \\ complexity for \\ large-scale applications}        &  2024     & \cite{PeakShaving}  \\[4pt]
				\hline
		Building          & \makecell[c]{Linear\\ model}       & \makecell[l]{Probabilistic \\ uncertainties \\ with chance \\ constraints}  & \makecell[c]{Stochastic \\ MPC} & \makecell[l]{Jointly account for  \\ the planning and\\ optimization of BHMES}                   & \makecell[l]{High computational \\ complexity;  Ignores \\ nonlinear characteristics\\ and safety constraints \\ of HESS }        &  2025     & \cite{liu2025coordinated}  \\[4pt]
		\hline
		Building        & \makecell[c]{Linear\\ model}        &  \makecell[l]{Learning from  \\ experience}    & \makecell[c]{Multi-agent \\  DRL}    & \makecell[l]{Considering building \\ thermal dynamics \\  for enhanced \\ operational flexibility }  &   \makecell[l]{Training instability  \\ and performance \\ sensitivity  to \\ hyperparameters}        &2022     & \cite{yu2022joint}  \\[4pt]
		\hline 
		\makecell[l]{Microgrids}    & \makecell[c]{Linear\\ model}        & \makecell[l]{Learning from  \\ experience}                    & \makecell[c]{Multi-agent \\  DRL}     & \makecell[l]{Considering energy \\ sharing  across   microgrids \\ for operating flexibility}       & \makecell[l]{Ignores nonlinear \\ characteristics and \\ safety constraints of HESS }  & 2025     & \cite{zhang2023multi}  \\[4pt]	
		\hline 
		Microgrid     
		& \makecell[c]{Nonlinear\\model}        
		& \makecell[l]{Learning from  \\ experience}    
		&  DRL                       
		& \makecell[l]{Incorporate  nonlinear \\ characteristics of CHP \\ and fuel cell}    & \makecell[l]{Depends on extensive\\ data or an appropriate \\ simulator. }           & 2023     & \cite{dolatabadi2022novel}  \\[4pt]
		\hline 
		\noalign{\vskip 2pt}
		\textbf{Community}
		& \makecell[c]{\textbf{Nonlinear}\\ \textbf{model}} 
		& \makecell[l]{\textbf{Learning from} \\ \textbf{experience}}     
		&  \makecell[c]{\textbf{SR}\\ \textbf{+ DRL}}
		& \makecell[l]{\textbf{Account for multi-physics} \\ \textbf{process and safe constraints}\\ \textbf{of HESS; Fast convergence} \\\textbf{and high-quality policy.}}  
		& \makecell[l]{\textbf{Depends on extensive}\\ \textbf{data or an appropriate} \\ \textbf{simulator}.}  
		&  \textbf{2026} 
		& \textbf{This paper}\\
		\bottomrule
	\end{tabular}
\end{table*}

\subsection{Paper contribution}
To address the existing gaps, this paper investigates  the optimal operation of an HMES that integrates renewables, HESS,  multiple types of energy storage and conversion devices with the goal of  jointly satisfying  multi-energy demands of a building community. \textbf{To enable  cost-efficient, reliable and intelligent operation of the HMES,  this paper makes the following main contributions:}
\begin{itemize}
    \item We establish a comprehensive model for the optimal operation of HMES that  incorporates the nonlinear operating characteristics as well as the multi-physics coupled process of HESS to ensure  reliable and efficient operation.

    \item  We propose a state-action representation learning enhanced deep reinforcement learning (SR-DRL) framework to  account for the system complexity,  which significantly improves learning convergence and performance over conventional DRL methods.

    \item  We further provide some insights into the role of the state–action representation learning (SR)  for DRL and conclude that it can reorganize the original state-space of HMES into a well-structured and cluster-aware presentation, thereby facilitating actor–critic learning in DRL methods.
   
\end{itemize}

We conduct a number of case studies in terms of the operation performance of an HMES  with real-world datasets. The results show that the proposed model can enable the safe operation of the system while the existing simplified  model may lead to safe constraints violation. 
Besides, the proposed learning framework can significantly  enhance the learning rate and quality of policy, decreasing both  operation cost and constraint violations under uncertainty.

The rest of the paper is as follows. Section~\ref{sec:modeling} introduces a multi-physics coupled model of HMES. Section~\ref{sec:SR-DRL} introduces the SR-DRL method for HMES Section~\ref{sec:Experiments and Results} presents the experimental results and analysis. Section~\ref{sec:conclusion} concludes this paper and discuss future works.

%% The second section
\section{Problem formulation}
\label{sec:modeling}

\subsection{Overview of HMES}
\begin{figure}[htbp]
	\centering
	\includegraphics[width=0.83\linewidth]{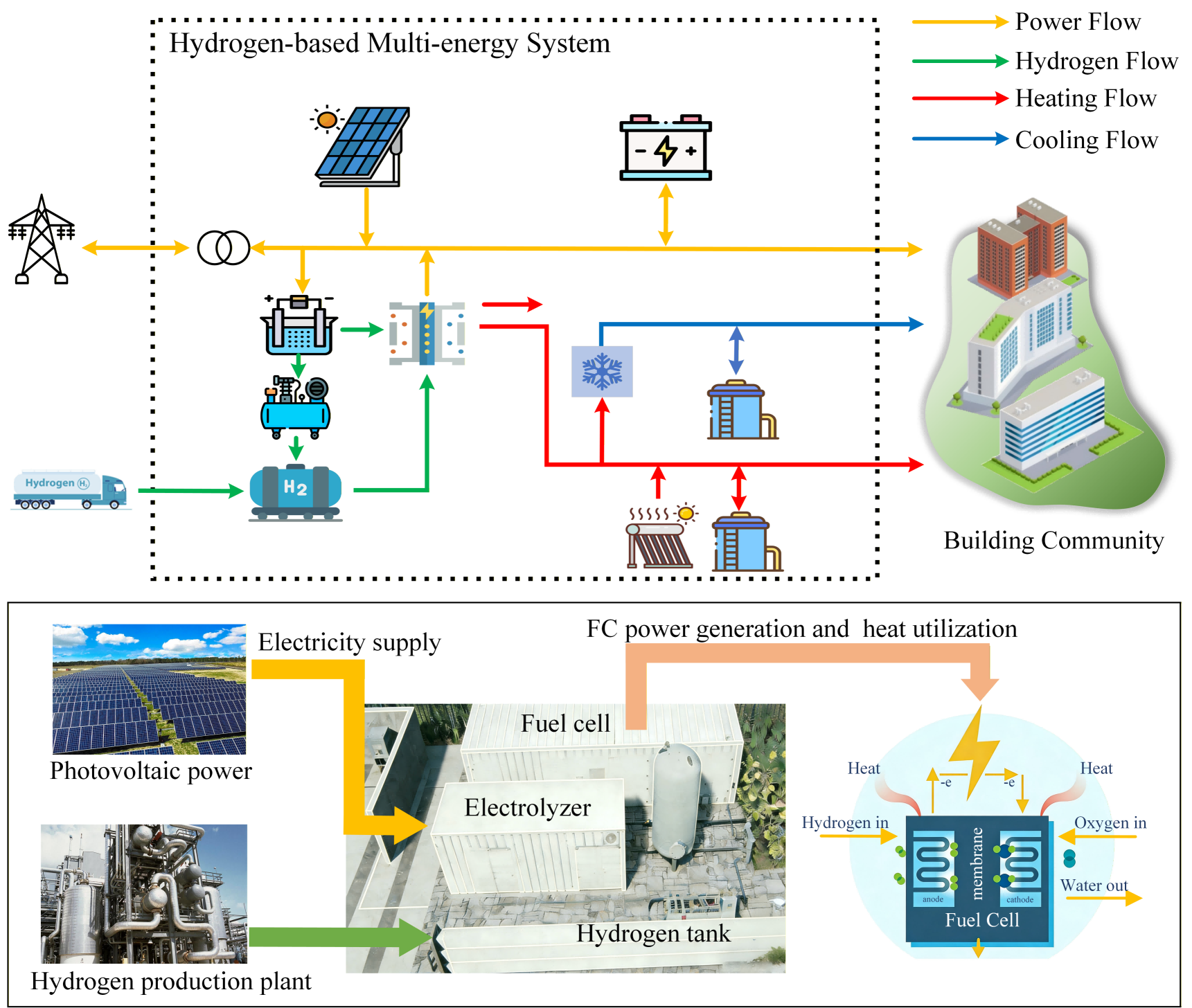} 
	\captionsetup{belowskip=-8pt}
	\caption{Hydrogen-based multi-energy system for a building community}
	% \vspace{-30pt}
	\label{fig:HIES}
\end{figure}

We consider the optimal operation of a hydrogen-based multi-energy system (HMES) integrated into a building community for supplying electricity, cooling and heating energy as  illustrated in Figure~\ref{fig:HIES}.
The system integrates electrical storage system, thermal and cooling energy storage (ESS, TES and CES) corresponding to batteries, hot and chilled water tanks, respectively.
Particularly, the system involves an advanced hydrogen energy storage system (HESS) consisting of electrolyzers (EL),  compressor unit, hydrogen tank, and fuel cells (FC), which enables the coordinated operation of multi-energy supply and demand. 
Hydrogen can be produced by renewable-powered water electrolysis and stored in the tank after cleaned and compressed by the compressor for later use. When needed, the stored hydrogen can be fueled by FC to generate electricity and heat.
A  heat recovery unit is coupled with the FC for capturing the residual heat, and then used  to heat water in the tank or converted to cooling by the absorption chiller (AC). 
The hot and chilled water tanks serve as thermal energy buffers for storing heating and cooling energy to satisfy diverse forms of heating and cooling demands. A solar thermal collector is also planted for the hot water tank to gain heat from solar radiations. 

This paper focuses on the optimal operation of the HMES to minimize the system's operating cost while satisfying the energy demands of the building community. We consider a discrete control framework with a decision interval of $\Delta t = 1h$ and each day equally divided into time slots $T: = \{1, 2, \cdots, 24\}$.

\textbf{Notations}: This paper uses the alphabets $P, g, q, v$ to represent electricity, heating, cooling and hydrogen  energy. The subscript \texttt{ces, tes, ces, hss} indicate the electrical, thermal, cooling and hydrogen storage, respectively.

\subsection{Energy Storage Devices} \label{sec:energy_storage}
In this section, we present mathematical models for different types of energy storage devices. 

\textbf{Electrical storage system (ESS) }:  The operating dynamics and physical limits of  ESS  can be modeled as 
\begin{subequations} \label{eq:ESS}
	\begin{align}
		&S_{t + 1}^{\rm ess} = S_t^{\rm ess}
		+ \Big(P_t^{\rm ess, ch} \eta^{\rm ess, ch}
		+ \frac{P_t^{\rm ess,  dis}}{\eta^{\rm ess,  dis}}\Big)\Delta t, 
		\label{eq:ess_energy_dynamics} \\
		&S^{\rm ess}_{\min} \le S_t^{\rm ess} \le S^{\rm ess}_{\max}, 
		\label{con:ess_energy_range} \\
		&-P^{\rm ess,  dis}_{\max} \le P_t^{\rm ess,  dis} \le 0,
		\label{con:ess_discharge_limits} \\
		&~~0 \le P_t^{\rm  ess, ch} \le P^{\rm ess,  ch}_{\max},
		\label{con:ess_charge_limits} \\
		&P_t^{\rm ess,  ch} \cdot P_t^{\rm ess,  dis} = 0,  \quad t \in T.
		\label{con:ess_complementarity}
	\end{align}
\end{subequations}
where  $S_t^{\rm ess}$ [kWh] represents the stored electricity at time $t$, and $P_t^{\rm ess,\rm ch}, P_t^{\rm ess,\rm dis}$ [kW]  are the charging and discharging power, with  $\eta^{\rm ess,ch}$ and $\eta^{\rm ces,  dis}$ denoting the corresponding efficiencies. Eq.~\eqref{eq:ess_energy_dynamics} models the evolution of stored electricity affected over the time.  
Cons.~\eqref{con:ess_energy_range} enforce the operating range of energy capacity of the storage device. Cons.~\eqref{con:ess_discharge_limits} and \eqref{con:ess_charge_limits} characterize the maximum charging and discharging power. Cons.~\eqref{con:ess_complementarity} are complementary constraints, enforcing non-simultaneous charging and discharging operation. 

\textbf{Thermal storage system (TES) }:  The thermal energy storage corresponds to  hot water tank. Their operating dynamics and physical limits are modeled as 
\begin{subequations}
	\begin{align}
		&S_{t + 1}^{\rm tes} = S_t^{\rm tes}
		+ \Big(g_t^{\rm tes, ch} \eta^{\rm tes, ch}
		+ \frac{g_t^{\rm tes,  dis}}{\eta^{\rm tes,  dis}}\Big)\Delta t, 
		\label{eq:tes_energy_dynamics} \\
		&S^{\rm tes}_{\min} \le S_t^{\rm tes} \le S^{\rm tes}_{\max}, 
		\label{con:tes_energy_range} \\
		&-g^{\rm tes,  dis}_{\max} \le g_t^{\rm tes,  dis} \le 0,
		\label{con:tes_discharge_limits} \\
		&~~0 \le g_t^{\rm  tes, ch} \le g^{\rm tes,  ch}_{\max},
		\label{con:tes_charge_limits} \\
		&g_t^{\rm tes,  ch} \cdot g_t^{\rm tes,  dis} = 0, \quad  t \in T.
		\label{con:tes_complementarity}
	\end{align}
\end{subequations}
where $S_t^{\rm tes}$ [kWh] represents the stored thermal energy in the  hot tank at time $k$, and $g_t^{\rm tes, ch}, g_t^{\rm tes, dis}$ [kW] are the charging and discharging rate, with  $\eta^{\rm tes, ch}$ and $\eta^{\rm tes, dis}$ denoting the corresponding efficiencies. Eq.~\eqref{eq:tes_energy_dynamics} models the dynamics of thermal energy stored in the hot water tank. Cons.~\eqref{con:tes_energy_range} enforces the permissible operating range of energy capacity of the storage device. Cons.~\eqref{con:tes_discharge_limits} and \eqref{con:tes_charge_limits}   characterize the maximum charging and discharging rates. Cons.~\eqref{con:tes_complementarity} enforces non-simultaneous charging and discharging of thermal energy storage.

\textbf{Cooling storage system (CES) }:  The cooling energy storage corresponds to the chilled water tank. Their operating dynamics and physical limits are modeled as 
\begin{subequations}
	\begin{align}
		&S_{t + 1}^{\rm ces} = S_t^{\rm ces}
		+ \Big(q_t^{\rm ces, ch} \eta^{\rm ces, ch}
		+ \frac{q_t^{\rm ces,  dis}}{\eta^{\rm tes,  dis}}\Big)\Delta t, 
		\label{eq:ces_energy_dynamics} \\
		&S^{\rm ces}_{\min} \le S_t^{\rm ces} \le S^{\rm ces}_{\max}, 
		\label{con:ces_energy_range} \\
		&-q^{\rm ces,  dis}_{\max} \le q_t^{\rm ces,  dis} \le 0,
		\label{con:ces_discharge_limits} \\
		&~~0 \le q_t^{\rm  ces, ch} \le q^{\rm ces,  ch}_{\max},
		\label{con:ces_charge_limits} \\
		&q_t^{\rm ces,  ch} \cdot q_t^{\rm ces,  dis} = 0, \quad  t \in T.
		\label{con:ces_complementarity}
	\end{align}
\end{subequations}
where $S_t^{\rm ces}$ [kWh] represents the stored cooling in the chilled water tank at time $k$, and $q_t^{\rm ces,  ch}, P_t^{\rm ces,  dis}$ [kW] are the charging and discharging rate, with  $\eta^{\rm ces,  ch}$ and $\eta^{\rm ces,  dis}$ denoting the corresponding efficiencies. Eq.~\eqref{eq:ces_energy_dynamics} models the dynamics of cooling energy stored in the tank. Cons.~\eqref{con:ces_energy_range} specifies the operating range of energy capacity of the storage. Cons.~\eqref{con:ces_discharge_limits} and \eqref{con:ces_charge_limits} characterize the maximum charging and discharging rates. Cons.~\eqref{con:ces_complementarity} ensures non-simultaneous charging and discharging operation of cooling energy storage. 

\subsection{Hydrogen energy storage system (HESS)}
\label{sec:HESS}
We assume the HESS consists of  a Siemens Silyzer 100 electrolyzer, a Swiss Hydrogen fuel cell, a compressor, and a pressurized hydrogen storage tank,  similar to \cite{PeakShaving}. The HESS suffers from pronounced nonlinear operating characteristics and admits a multi-physics coupled process. This paper adopts the formulations  from \cite{flamm2021electrolyzer, PeakShaving}. 

\textbf{Electrolyzers (EL)}:  The Siemens SILYZER 100 electrolyzer  is a self-contained system consisting of four stacks, each of which has a nominal rated power of $25$ kW and  a total power of $100$ kW. The stacks can be run at an overloaded power of $50$ kW for a limited period of time (about 15 mins).   To ensure safe and efficient operation of EL,  stack temperature must be maintained below a prescribed upper limit. A cooling device with a programmed control logic is often coupled with EL for thermal management.  
For example, the cooling device is set to  on and off  when the stack temperature exceeds
$50$°C or drops below $40$°C for EL for maintaining the target stack temperature of $60$°C in  \cite{flamm2021electrolyzer}.
The control input of EL is the injected DC power $P^{\rm ely}_t$ in the range of $[P^{\rm ely}_{\min}, P^{\rm ely}_{\max}] = [15, 200]$ kW with nominal power of $P^{\rm ely}_{\rm nom} = 100$ kW. The energy conversion efficiency of EL depends on the ranges of DC power injections and stack temperature, exhibiting a nonlinear operating characteristic, which can be captured by  piece-wise affine function
\begin{equation} \label{eq:v_p}
	v^{\rm{ely}}_t = \begin{cases} 
		z_1 T^{\rm{ely}}_t + z_0 + z_{{\rm{low}}}(P^{\rm{ely}}_t -P^{{\rm{ely}}}_{\rm{nom}}) & \text{if } P^{\rm{ely}}_{\min} < P^{\rm{ely}}_t \leq P^{{\rm{ely}}}_{\rm{nom}}, \\
		z_1 T^{\rm{ely}}_t + z_0 + z_{{\rm{high}}}(P^{\rm{ely}}_t - P^{{\rm{ely}}}_{\rm{nom}}) & \text{if }P^{{\rm{ely}}}_{\rm{nom}} < P^{\rm{ely}}_t \leq P^{\rm{ely}}_{\max}, \\
		0 & \text{otherwise.}
	\end{cases}
\end{equation}
where $v^{\rm{ely}}_t$ [$\rm m^3/h$] denotes the volume rate of produced hydrogen. 
$P^{\rm{ely}}_t$ and $T^{\rm{ely}}_t$ are the injected DC power  and stack temperature of EL. The model captures the power-dependent energy efficiency of the EL using parameters $z_{\rm low}$ and $z_{\rm high}$. Specifically, the EL exhibits higher energy conversion efficiency at lower injected power levels, with $z_{\rm low}$ typically slightly exceeding $z_{\rm high}$.

The stack temperature of EL are governed by internal heat generation rate $\dot{Q}_{\rm gen}$, heat dissipation rate $\dot{Q}_{\rm loss}$ and the cooling power $\dot{Q}_{\rm cool}$  supplied by cooling device. This leads to the following lumped-parameter thermal balance model
\begin{equation}
\begin{split}
	\dfrac{dT^{\rm ely}_t}{dt} = K_1 \dot{Q}_{\rm gen} - K_2 \dot{Q}_{\rm loss} - K_3 \dot{Q}_{\rm cool}
\end{split}
\end{equation} 

We assume  a similar  on-off  control logic  of  cooling device in \cite{flamm2021electrolyzer} and the stack temperature dynamics can be presented in the compact form
\begin{equation} \label{eq:EL_T}
	T^{\rm{ely}}_t = {j_1}T^{\rm{ely}}_{t-1} + {j_2}P^{\rm{ely}}_{t-1} + {j_0}, ~~\forall t \in T. 
\end{equation}
where the effects of heat generation, loss and  cooling devices are encapsulated in  model parameters $j_0, j_1, j_2$. The DC power injections $P^{\rm{ely}}_t$ is involved to capture the heat generation and loss effects under different operating conditions.

To ensure safe and reliable operation of EL,  stack temperature must be maintained below a upper limit $T^{\rm ely}_{\max}$ (often 70°C), which can be modeled as 
\begin{equation} \label{con:EL_T}
0 \leq	T^{\rm{ely}}_t  \leq T^{\rm ely}_{\max}, ~~ \forall t \in T. 
\end{equation}

As introduced,  though the stacks of ES have a nominal power of 25kW, they can  work at power up to $50$ kW for certain short time periods (i.e., $15$ minutes). This behavior can be captured by an overload counter that records the number of time slots that the stack current  overloads. Specially, for a EL with four stacks, the injected DC power will be  first allocated  to Stack $1$   and then the other three.  Therefore, the first stack tends to overload first and can be used as an indicator. For the first stack, the allocated DC power $P^{\rm{ely}}_{t, 1}$ from the total DC power injections  can ba approximated by the following PWL function
\begin{equation} \label{eq:P_P1}
	P^{\rm{ely}}_{t, 1} = \begin{cases}
		P^{\rm{ely}}_t 
		& \text{if } 
		P^{\rm ely}_{\min} < P^{\rm{ely}}_t \leq {P^{{\rm{ely}}}_{\rm{nom}}}/{4},  \\
		
		{P^{{\rm{ely}}}_{\rm{nom}}}/{4} 
		& \text{if } 
	{P^{{\rm{ely}}}_{\rm{nom}}}/{4} < P^{\rm{ely}}_t \leq P^{{\rm{ely}}}_{\rm{nom}},  \\
		
		{P^{\rm{ely}}_t}/{4} 
		& \text{if }
		P^{{\rm{ely}}}_{\rm{nom}} < P^{\rm{ely}}_t \leq P^{\rm{ely}}_{\max}, \\
		0 & \text{otherwise.}
	\end{cases}
\end{equation}

The stack current depends on  the allocated power and stack temperature, which can be captured by 
\begin{equation} \label{eq:P1_i}
	i^{\rm{ely}}_{t} = \begin{cases}
		h_1 T^{\rm{ely}}_t + h_0 + h_{{\rm{low}}}(P^{\rm{ely}}_{t, 1} - P^{\rm{ely}}_{ \rm{nom}, 1}) & \text{if } 
		P^{\rm{ely}}_{\min} <P^{\rm{ely}}_{t, 1} \leq  P^{\rm{ely}}_{ \rm{nom}, 1}, \\
		h_1 T^{\rm{ely}}_t + h_0 + h_{{\rm{high}}}(P^{\rm{ely}}_{t, 1} -  P^{\rm{ely}}_{ \rm{nom}, 1})
		& \text{if }  
		P^{\rm{ely}}_{ \rm{nom}, 1} <P^{\rm{ely}}_{t, 1}\leq P^{\rm{ely}}_{\max}, \\
		0 & \text{otherwise.}
	\end{cases}
\end{equation}
where $P^{\rm{ely}}_{ \rm{nom}, 1} = 25$ kW denotes the nominal power of stack $1$.  The nonlinear characteristics are captured by model parameters $h_0, h_1, h_{\rm low}, h_{\rm high}$. 

A overload counter that tacks the current of stack $1$ for ensuring the safe operation of EL can be modeled as 
 \begin{subequations}
 	\begin{align}
& C^{\rm ely}_t = \max\left\{0, C^{\rm ely}_{t-1} + (i^{\rm ely}_{t} - i^{\rm ely}_{\rm nom})\cdot \Delta t \right\},     \label{eq:i_C} \\
& C^{\rm ely}_t \leq C^{\rm ely}_{\max}, \quad  \forall t \in T.  \label{con:EL_current}
 	\end{align}
 \end{subequations} 
where $i^{\rm ely}_{\rm nom} = 300~A$ denotes the nominal stack current. 
$C^{\max}$ denotes the overload counter upper limits determined by the permissible overload time periods of  EL.  

\textbf{Hydrogen tank}: Hydrogen produced by EL is cleaned, purified and compressed and then storage in the hydrogen tank. The stored hydrogen is consumed by FC to generate electricity and heat.  The safe and efficient operation of hydrogen storage should account for the tank capacity, temperature and pressure limits, which can be modeled as 
\begin{subequations}
\begin{align}
	&S_t^{\rm hss} = S_{t - 1}^{\rm hss} + \left(\alpha  \cdot v^{\rm{ely}}_t - v^{\rm{fc}}_t  + v_t^{\rm buy}\right) \Delta t, \label{eq:hydrogen_dynamics} \\
	&S^{\rm hss}_{\min} \le S_t^{\rm hss} \le S^{\rm hss}_{\max}, 
	\label{con:hss_energy_range} \\
	&T^{\rm{tank}}_t = {g_0}T^{\rm{tank}}_{t-1} + {g_1}T^{\rm amb}_{t-1}, \label{eq:tank_temperature_dynamics}  \\ 
	&p^{\rm tank}_t = ({b_0} + {b_1} T^{\rm{tank}}_t) \cdot {S^{\rm hss}_t \cdot {\rho _0}}/{V^{\rm tank}}, \label{eq:tank_pressure_dynamics}\\ 
	& -v^{\rm buy}_{\max} \leq v_t^{\rm buy} \leq v^{\rm buy}_{\max},  \label{eq:hydrogen_trading}\\
	&p^{\rm tank}_{\rm min} \leq p^{\rm tank}_t \leq p^{\rm tank}_{\rm max}, \quad \forall t \in T.  \label{con:tank_pressure_limits} 
\end{align}
\end{subequations}
where $S^{\rm hss}_t$ [$\rm m^3$] represents the volume of stored hydrogen at time $t$ (under standard atmospheric pressure). $T_t^{\rm tank}$ and  $p_t^{\rm tank}$  characterize the tank temperate and pressure. Eq.~\eqref{eq:hydrogen_dynamics} models the dynamics of hydrogen stored in the tank influenced by the volume rate of produced hydrogen $v^{\rm ely}_t$ by EL, consumed hydrogen $v^{\rm fc}_t$ by FC and net purchase of hydrogen $v_t^{\rm buy}$ in market.  
The relatively small non-negative constant  $\alpha \in [0, 1]$ is involved to model the  loss of hydrogen through compressor. 
The storage capacity of hydrogen tank is implicitly determined by its pressure, which is affected by the stored hydrogen and tank temperature. Cons.~\eqref{con:hss_energy_range} capture the safe and permissible operating volume range of the hydrogen tank.
Eq.~\eqref{eq:tank_temperature_dynamics} capture the temperature dynamics of the tank which is affected by tank thermal inertia  and heat exchange with the ambient environment captured by model parameters $g_0$ and $g_1$ with  $T^{\rm amb}_t$ denoting the  ambient temperature. Eq.~\eqref{eq:tank_pressure_dynamics} evaluate the tank pressure which is affected by tank temperature and the volume of stored hydrogen with $V^{\rm tank}$ denoting the volume of  tank.  
Cons.~\eqref{eq:hydrogen_trading} enforces the limits of transaction in hydrogen market.  
To ensure safe operation, the tank pressure must be maintained in $[0, 40]$ bar as captured by Cons.~\eqref{con:tank_pressure_limits}.

\textbf{Full cells (FC)}:  We consider a Proton Exchange Membrane Fuel Cell (PEMFC) comprising two stacks each with a maximum power of $50$ kW and $100$ kW in total. 
FC converts hydrogen $v_t^{\rm fc}$ into stack current $i_t^{\rm fc}$. The resulting stack current produces DC power $P_t^{\rm fc}$. This process exhibits nonlinear behavior and can be modeled as
\begin{subequations}
	\begin{align}
		& v_t^{\rm fc} = c \cdot i^{\rm fc}_t,  \label{eq:v_i} \\
		& i_t^{\rm fc} = 
		\begin{cases}
			s_1 P_t^{\rm fc}, & \text{if } P^{\rm fc}_{\min} < P_t^{\rm fc} \le P^{\rm fc}_{\rm bp}, \\
			s_2 (P_t^{\rm fc} - P^{\rm fc}_{\rm bp}) + i^{\rm fc}_{\rm bp}, & \text{if } P^{\rm fc}_{\rm bp} < P_t^{\rm fc} \le P^{\rm fc}_{\max}, \\
			0, & \text{otherwise},
		\end{cases} \label{eq:P_i}\\
		& P^{\rm fc}_{\min} \leq P^{\rm fc}_t \leq P^{\rm fc}_{\max}, \quad \forall t \in T. \label{con:DC_power_limits}
	\end{align}
\end{subequations}
where $v_t^{\rm fc}$ denotes the volume rate of hydrogen injected into the FC. $c$ is an experimental constant accounting for
purging and Faradaic losses. The conversion of DC power to stack current shows a nonlinear characteristics as captured by the PWL function \eqref{eq:P_i} with $(P^{\rm fc}_{\rm bp}, i^{\rm fc}_{\rm bp})$  denoting the breakpoints and $s_1, s_2$ as energy conversion efficiencies. Cons.~\eqref{con:DC_power_limits} characterize the range of the DC output of FC.  

While generating electricity, the FC produces heat that are captured by the heat recover unit
\cite{assafTransientSimulationModelling2016}
\begin{equation} \label{eq:g_u}
{g^{\rm fc}_t} = \eta _{\rm rec}^{\rm fc} \cdot (1 - \eta ^{\rm fc }) /\eta ^{ \rm fc } \cdot P^{\rm fc}_t, \forall t \in T. 
\end{equation}
where $\eta _{{\rm{rec}}}$ denotes the heat recovery efficiency and $\eta ^{{\rm{fc}}}$ represents the heat conversion efficiency of the FC.

\subsection{Energy Conversion Devices}

\textbf{Absorption Chiller (AC)}:  The AC can convert heating energy to cooling energy. 
The produced cooling power of AC is determined by the injected heating power $g_t^{\rm ac}$ and energy conversion efficiency $\eta^{\rm ac}$, while respecting the operating limits. These can be modeled as  
\begin{equation}
	\begin{split}
		&q^{\rm ac}_t = g^{\rm ac}_t \cdot \eta^{\rm ac},  \\
		& g^{\rm ac}_t \leq g^{\rm ac}_{\max}, \quad t \in T. 
	\end{split}
\end{equation}
where $g^{\rm ac}_t$ denotes the injected heating power to AC. 
$q_t^{\rm ac}$ denotes the produced cooling energy by the AC. $g^{\rm ac}_{\max}$ captures the working limits of the AC. 

\textbf{Photovoltaic power (PV)}: Solar power generation is determined by the incident solar irradiance and the effective panel area. The following model is commonly used to characterize PV output:
\begin{equation}
	P_t^{\rm solar} = \eta^{\rm pv} \cdot A_{\rm pv} \cdot Q_t^{\rm rad}, \quad \forall t \in T. 
\end{equation}
where $P_t^{\rm solar}$ [kW] denotes the solar power output at time $t$, which is determined by the PV conversion efficiency $\eta^{\rm pv}$, the effective solar collection area $A_{\rm pv}$ [m$^2$], and the incident solar radiation  $Q_t^{\rm rad}$ [kW/m$^2$].

\textbf{Solar Thermal Connector}:  The solar thermal collector absorbs solar radiation to generate thermal energy for heating water. The thermal energy injected into the hot water tank can be modeled as
\begin{equation}
	g_t^{\rm solar} = \eta^{\rm stc} \cdot A_{\rm stc} \cdot Q_t^{\rm rad}, \quad \forall t \in T. 
\end{equation}
where $g_t^{\rm solar}$ [kW] denotes the absorbed heat by solar thermal collector which depends on the energy conversion efficiency $\eta^{\rm stc}$,  the collector area $A_{\rm stc}$  [$m^2$] and  incident solar radiation $Q_t^{\rm rad}$  [kW/m$^2$].

\subsection{Energy Balances}
The fundamental function of the HMES is to ensure the supply and demand balance at each time slot. This can be captured by the following multi-energy (i.e., electricity, heating and cooling) balance equations
\begin{subequations} \label{eq:energy_balance}
\begin{align}
	&P_t^{\rm g} + P_t^{\rm solar} +  P_t^{\rm fc}= P_t^{\rm ess,ch} + P_t^{\rm ess,dis} +  P_t^{\rm ely} + Q_t^{\rm ED},  \label{eq:power_balance} \\
	&g_t^{\rm solar} + g_t^{\rm fc} = g_t^{\rm tes,ch} + g_t^{\rm tes,dis} + g_t^{\rm ac} + Q_t^{\rm HD},  \label{eq:heat_balance}\\
	&q_t^{\rm ac} = q_t^{\rm ces,ch} + q_t^{\rm ces,dis} + Q_t^{\rm CD}, \quad \forall t \in T. \label{eq:cool_balance}
\end{align}
\end{subequations}
where $P^{\rm g}_t$ denotes the purchase electricity from electricity market. $P^{\rm solar}_t$ denotes the on-site PV generation.  $Q_t^{\rm ED}, Q_t^{\rm HD}$, $Q_t^{\rm CD}$ denote the electricity, heating, and cooling demand of building community. 

\subsection{Objective function}
Our objective is to minimize the operating cost of HMES for supplying  multiple energy demands of the building community. 
The daily operating cost includes electricity and hydrogen trading costs
\begin{equation}
	J_{\rm cost} \!\!=\! \mathbb{E} \left\{\sum_{ t \in T} \left[\frac{\lambda^{\rm b}_t \!-\! \lambda^{\rm s}_t}{2} \lvert P^{\rm g}_t \rvert + \frac{\lambda^{\rm b}_t  + \lambda^{\rm s}_t}{2}  P^{\rm g}_t   + \lambda^{\rm h}_t v^{\rm buy}_t  \right]\Delta t \right\}
	\label{eq:objective}
\end{equation}
where $\lambda^{\rm b}_t$ and $\lambda^{\rm s}_t$ are  the buying and selling prices in the electricity market at time $t$. $\lambda^{h}_t$ denotes the  hydrogen trading price (i.e., both purchasing and selling). 
The first two terms are electricity cost, and the third is hydrogen cost.

Overall, the optimal operation of the HMES corresponds to solving the stochastic optimization problem: $\{\min J_{\rm cost} ~{\rm s.t.}~ \eqref{eq:ESS} - \eqref{eq:energy_balance}\}$ which suffers high computational complexity from extensive decision variables, nonlinear constraints, and multiple sources of uncertainties.

\section{Representation Learning enhanced DRL} \label{sec:SR-DRL}

To address the aforementioned challenges, this paper develops an advanced deep reinforcement learning (DRL)  method to enable effecient and optimal operating of the HMES. 
Specifically, historical data of market prices, renewable output, and multiple energy demands are used to train a DRL agent, which is then deployed for real-time operation. During training, a multi-physics coupled model is employed as a simulator, enabling the DRL agent to acquire extensive operational experience. 
Whereas the method is model-free and does not rely on explicit mathematical formulations if a physical simulator or practical system is allowed to interact.

\subsection{Markov Decision Process (MDP)}  
To develop a DRL-based approach, the problem is first reformulated as a Markov Decision Process (MDP), which is a general framework for sequential decision-making under uncertainty. An MDP is formally defined by a 5-tuple $(\mathcal{S}, \mathcal{A}, \mathcal{P}, r, \gamma)$, including the state space $\mathcal{S}$, action space $\mathcal{A}$, state transition probability $ \mathcal{P}$, reward function $r$ and discounting factor $\gamma$.
In the following, we present the MDP reformulation for the optimal operation of HMES based on the mathematical model of previous section.

\textbf{State}: The state should involve all required information to make decision at each stage. For the optimal operation of the HMES, we define the system state at time $t$ as 
\begin{equation}
	s_t =
	\left(
	\begin{aligned}
		& day,\, hour,\, \lambda^{\rm b}_{t},\, \lambda^{\rm s}_{t},\, \lambda^{\rm h}_{t},\, Q^{\rm rad}_{t}, \\
		& S^{\rm ess}_{t},\, S^{\rm tes}_{t},\, S^{\rm ces}_{t},\, S^{\rm hss}_{t},\,
		T^{\rm ely}_t,\, C^{\rm ely}_t,\, T^{\rm tank}_t, \\
		& Q^{\rm ED}_{t},\, Q^{\rm HD}_{t},\, Q^{\rm CD}_{t}
	\end{aligned}
	\right)
\end{equation}
where $day \in \{1, 2, \cdots, 7\}$  denotes the day of the weak and $hour \in \{1, 2, \cdots, 24\}$ denotes the hour. $\lambda^{\rm b}_{t},\, \lambda^{\rm s}_{t},\, \lambda^{\rm h}_{t}$ are electricity and hydrogen market price. $Q^{\rm rad}_{t}$ is incident solar radiation for PV cells and solar thermal collectors. $S^{\rm ess}_{t}$, $S^{\rm tes}_{t}$, $S^{\rm ces}_{t}$, and $S^{\rm hss}_{t}$ denote the energy levels of electrical, thermal, cooling, and hydrogen storage systems, respectively.
$T^{\rm ely}_t,\, C^{\rm ely}_t,\, T^{\rm tank}_t$ are stack temperature, overload counter of EL, and temperature of hydrogen tank of HESS. $Q^{\rm ED}_{t},\, Q^{\rm HD}_{t},\, Q^{\rm CD}_{t}$ are electricity, heating and cooling demands of the building community. 
Some studies incorporate historical information, e.g., $\lambda^{\rm b}_{t-k:t},\, \lambda^{\rm s}_{t-k:t},\, \lambda^{\rm h}_{t-k:t}, Q^{\rm ED}_{t-k:t},\, Q^{\rm HD}_{t-k:t},\, Q^{\rm CD}_{t-k:t}$, to enhance learning performance. This is avoided in this paper as they increase state dimensionality without performance gain.

\textbf{Action}: The action $a_t \in A$ specifies the scheduling decisions of HMES at time  $t$ and is defined as   
\begin{equation} \label{eq:action}
	a_t = \left( a^{\rm ess}_t,  a^{\rm tes}_t, a^{\rm css}_t, a^{\rm hss}_t, a^{\rm buy}_t \right)
\end{equation}
where $a^{\rm ess}_t,  a^{\rm tes}_t, a^{\rm css}_t, a^{\rm hss}_t \in [-1,1]$ denote the normalized charging and discharging  of electrical, thermal, cooling, and hydrogen storage. 
$a^{\rm buy}_t \in [-1,1]$ indicates normalized transaction in hydrogen market. A positive value indicates charging or buying, while a negative value denotes discharging or selling. Note that not all control variables of mathematical models are included as action, as the remaining variables can be derived from system equations.
While executing, the normalized action should be converted to practical device operations considering their physical limits as follows.

For electrical, thermal, cooling and hydrogen storage, the actual charging and discharging, considering physical limits, are given by
\begin{equation}
	\begin{split}
		&P^{\rm ess, ch}_t = {\rm clip} \left(a^{\rm ess}_t P^{\rm ess,  ch}_{\max}, ~~~~0,  ~~~~\frac{S^{\rm ess}_{\max} - S^{\rm ess}_t}{\eta^{\rm ess,  ch} \Delta t}\right),\\
		&P^{\rm ess,  dis}_t \!=\! {\rm clip} \left(a^{\rm ess}_t P^{\rm ess,  dis}_{\max}, \frac{( S^{\rm ess}_{\min} -S^{\rm ess}_t) \eta^{\rm ess, dis}}{ \Delta t}, 0\right), \forall t \in T. 
	\end{split}
\end{equation}
\begin{equation}
	\begin{split}
		&g^{\rm tes, ch}_t = {\rm clip} \left( a^{\rm tes}_t g^{\rm tes,  ch}_{\max}, ~~~~0,  ~~~~\frac{S^{\rm tes}_{\max} - S^{\rm tes}_t}{\eta^{\rm tes,  ch} \Delta t}\right),\\
		&g^{\rm tes,  dis}_t \!=\! {\rm clip} \left(a^{\rm tes}_t P^{\rm tes,  dis}_{\max}, \frac{( S^{\rm tes}_{\min} -S^{\rm tes}_t) \eta^{\rm tes, dis}}{ \Delta t}, 0\right), \forall t \in T. 
	\end{split}
\end{equation}
\begin{equation}
	\begin{split}
		&q^{\rm ces, ch}_t = {\rm clip} \left( a^{\rm ces}_t q^{\rm ces,  ch}_{\max}, ~~~~0,  ~~~~\frac{S^{\rm ces}_{\max} - S^{\rm ces}_t}{\eta^{\rm ces,  ch} \Delta t}\right),\\
		&q^{\rm ces,  dis}_t \!=\! {\rm clip} \left(a^{\rm ces}_t q^{\rm ces,  dis}_{\max}, \frac{( S^{\rm ces}_{\min} -S^{\rm ces}_t) \eta^{\rm ces, dis}}{ \Delta t}, 0\right), \forall t \in T. 
	\end{split}
\end{equation}
\begin{equation}
	\begin{split}
		&v^{\rm hss, ch}_t = {\rm clip} \left( a^{\rm hss}_t v^{\rm hss,  ch}_{\max}, ~~~~0,  ~~~~\frac{S^{\rm hss}_{\max} - S^{\rm hss}_t - v^{\rm buy}_t }{\Delta t}\right),~~~\\
		&v^{\rm hss,  dis}_t \!=\! {\rm clip} \left(a^{\rm hss}_t v^{\rm hss,  dis}_{\max}, \frac{S^{\rm hss}_{\min}\!-S^{\rm hss}_t - v^{\rm buy}_t}{ \Delta t}, ~0\right), \forall t \in T.~~~ 
	\end{split}
\end{equation}
where ${\rm clip}(x, y, z) \triangleq \min\big(\max(x, y), z\big)$ bounds $x$ within $[y, z]$. $v^{\rm hss,  dis}_{\max}, v^{\rm hss,  ch}_{\max}$ denote the maximum charging and discharging rate of hydrogen tank, which can be determined based on experience. 
This ensures that electrical, thermal and cooling storage are not overcharged or over-discharged during operation.

The actual net purchase of hydrogen in  market is 
 \begin{equation}
 	\begin{split}
      v_t^{\rm buy} = a_t^{\rm buy} v^{\rm buy}_{\rm max}, \forall t \in T. 
	\end{split}
\end{equation}

The other control variables of HMES can be derived from the system equations. Specifically, the control inputs of AC can be determined by the  energy balance equations: 
 \begin{equation}
	\begin{split}
		& q_t^{\rm ac} = q_t^{\rm ces,ch} + q_t^{\rm ces,dis} + Q_t^{\rm CD}, \\
		& g_t^{\rm ac} = q_t^{\rm ac}/\eta^{\rm ac}, \forall t \in T. 
	\end{split}
\end{equation}

Further, the working status of FC can be determined by
 \begin{equation}
	\begin{split}
		& g_t^{\rm ac} \xrightarrow{\eqref{eq:heat_balance}}  g_t^{\rm fc}  \xrightarrow{\eqref{eq:g_u}} P_t^{\rm fc} \xrightarrow{\eqref{eq:P_i}} i_t^{\rm fc} \xrightarrow{\eqref{eq:v_i}} v_t^{\rm fc} 
	\end{split}
\end{equation}

Given the purchased hydrogen $v_t^{\rm buy}$, the hydrogen consumption of FCs $v_t^{\rm fc}$, and the net charging and discharging volumes $v_t^{\rm hss, ch}, v_t^{\rm hss, dis}$, we can obtain $v_t^{\rm ely}$ from the following hydrogen balance equation
 \begin{equation}
	\begin{split}
		v_t^{\rm hss, ch} + v_t^{\rm hss, dis} = \alpha \cdot  v_t^{\rm ely} - v_t^{\rm fc} + v_t^{\rm buy}, \forall t \in T. 
	\end{split}
\end{equation}	

We can further determine the operating status of EL as 
 \begin{equation}
	\begin{split}
v_t^{\rm ely} \xrightarrow{\eqref{eq:v_p}} P_t^{\rm ely} \xrightarrow{\eqref{eq:EL_T}} T_t^{\rm ely}  \xrightarrow{\eqref{eq:P_P1}} P^{\rm{ely}}_{t, 1} \xrightarrow{\eqref{eq:P1_i}}  i^{\rm{ely}}_{t} \xrightarrow{\eqref{eq:i_C}}   C^{\rm{ely}}_{t}  
	\end{split}
\end{equation}

Additionally, the amount of electricity $P_t^{\rm g}$ purchased from the electricity market is determined by the electricity balance \eqref{eq:power_balance}.

\textbf{State transition}: The state transitions describe the evolution of states under given control inputs. The time components:  \emph{day} and \emph{hour} are deterministic and follow the calendar. The external states: market prices
$\lambda^{\rm b}_{t}, \lambda^{\rm s}_{t},  \lambda^{\rm h}_{t}$, solar radiation $Q_t^{\rm rad}$ and energy demands $Q^{\rm ED}_{t},\, Q^{\rm HD}_{t},\, Q^{\rm CD}_{t}$ are not explicitly available and are captured by historical data. The storage levels $S^{\rm ess}_{t},\, S^{\rm tes}_{t},\, S^{\rm ces}_{t},\, S^{\rm hss}_{t}$, stack temperature, overload counter and tank temperature  $T^{\rm ely}_t,\, C^{\rm ely}_t,\, T^{\rm tank}_t$ follow the system dynamics described in Section \ref{sec:energy_storage} and \ref{sec:HESS}. 

\textbf{Reward}:  
An appropriate reward should be designed to guide the learning process of  DRL agent. Our objective is to minimize the operating cost for satisfying the multiple energy demands while considering physical operating limits. 
We therefore define the reward as the negative operating cost at each time slot. Considering the limitation of DRL in handling constraints,
we also involve the hard constraints of HESS in the objective as penalty.  
The reward function is defined as 
\begin{equation*}
	\begin{split}
		&r_t = - \left[\frac{\lambda^{\rm b}_t - \lambda^{\rm s}_t}{2}  \lvert P^{\rm g}_t \rvert + \frac{\lambda^{\rm b}_t   + \lambda^{\rm s}_t}{2}  P^{\rm g}_t   + \lambda^{\rm h}_t  v^{\rm buy}_t  \right]\Delta t - \lambda \cdot penalty \\
		&penalty = ~~\left[C^{\rm ely}_t - C^{\rm ely}_{\max}\right]_{+}  +  \left[T^{\rm ely}_t - T^{\rm ely}_{\max}\right]_{+}\\
		& ~~\qquad\quad\quad + \left[ p^{\rm  tank}_{\min} - p^{\rm tank}_t \right]_{+}  + \left[p^{\rm tank}_t - p^{\rm  tank}_{\max}\right]_{+} \\
		%& {\rm device\_over\_limit} =
		& ~~\qquad\quad\quad + \left[P^{\rm fc}_{\min} - P_t^{\rm fc} \right]_{+} +  \left[P_t^{\rm fc} - P^{\rm fc}_{\max} \right]_{+}	 \\
		& ~~\qquad\quad\quad + \left[P^{\rm ely}_{\min} - P_t^{\rm ely} \right]_{+} +  \left[ P_t^{\rm ely} - P^{\rm ely}_{\max} \right]_{+} + \left[ g_t^{\rm ac} -  g^{\rm ac}_{\max}\right]_{+} \\
	\end{split}
\end{equation*}
where $\lambda \geq 0 $ is a non-negative penalty factor to trading off operating cost and constraints penalties, which can be tuned by experience. 

\begin{figure}[htbp]
	\centering
	\includegraphics[width=0.9\linewidth]{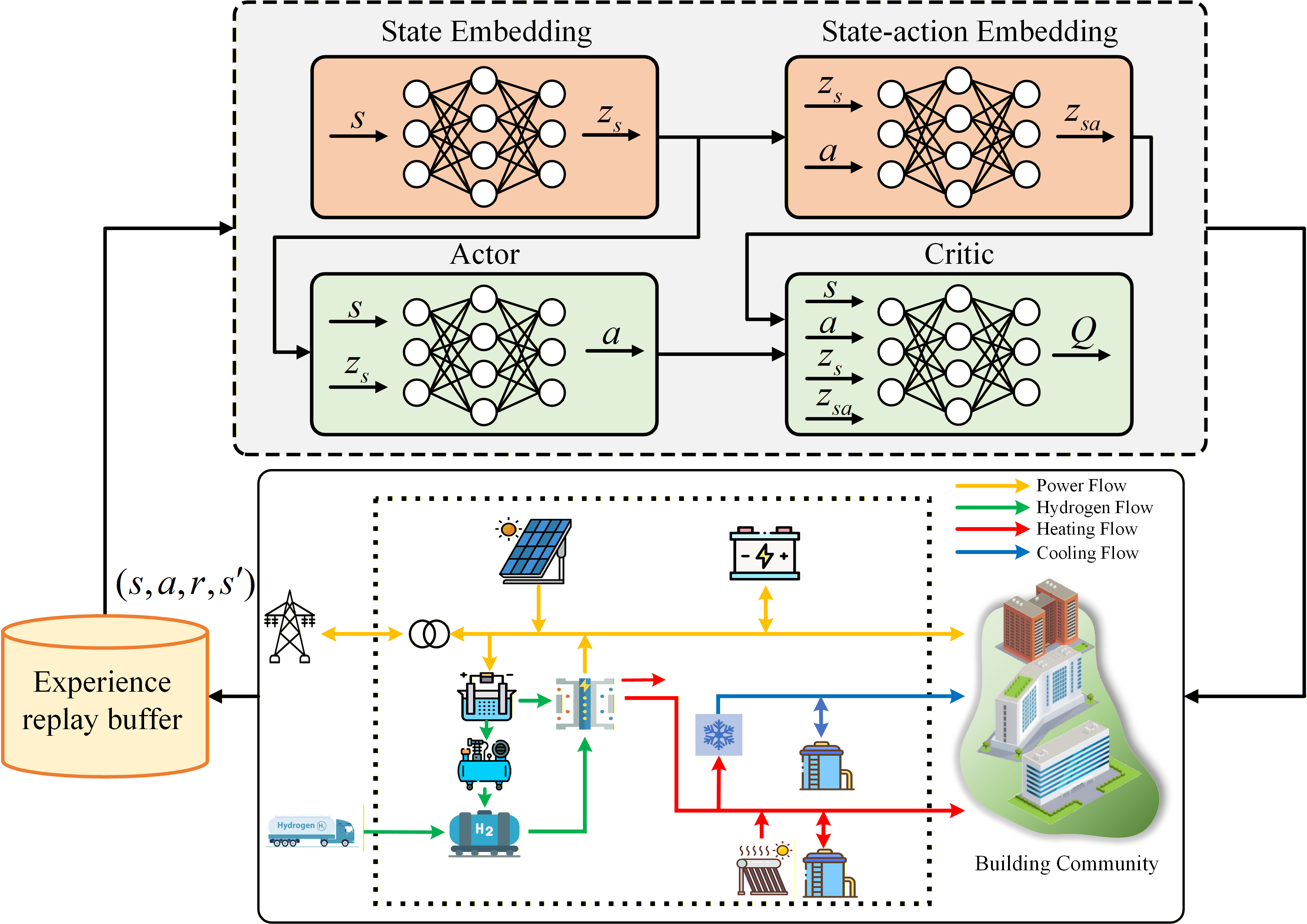}
	\caption{The architecture of representation learning enhanced DRL method}
	\label{fig:SR-DRL}
\end{figure}

\subsection{State-Action Reinforcement Learning for DRL} \label{sec:sr-drl}

Motivated by the idea of~\cite{fujimoto2023sale}, this section proposes a representation learning enhanced deep reinforcement learning (SR-DRL) framework for complex multi-energy systems  involving strong temporal and spatial couplings. 
The key idea of the SR-DRL method is to involve the state and state-action representing learning into the DRL pipeline. 
Specifically, \texttt{State Embedding} and \texttt{State-action Embedding} modules are integrated to augment the inputs of the actor and critic networks. These modules, referred to as \texttt{Representation Learning} in this paper. 
We emphasize that many existing works have explored  representation learning for state dimension reduction or abstraction in vision-based learning tasks~\cite{Learning-invariant-representations, Stochasti-latent-actor-critic, Deep-spatial-autoencoders, Target-driven-visual-navigation}, whereas the proposed method leverages representation learning to extract features that improve learning performance. 
The details of the state and state-action embedding are presented below.

\textbf{State and State-action Embedding}:
We define the \texttt{state embedding} as a mapping $\varPhi: \mathcal{S} \to \mathcal{Z}_s$, which transforms a raw state $s$ into a latent representation $z_s$.
The \texttt{state-action embedding} is defined  as $\varPsi: (\mathcal{S}, \mathcal{Z}_s) \to \mathcal{Z}_{sa}$, which maps the latent state-action pair $(z_s, a)$ to an embedding $z_{sa}$. 
Accordingly, the embeddings are given by
\begin{equation}
\begin{split}
		& z_s = \varPhi(s)   \\
		& z_{sa} = \varPsi(z_s, a)
\end{split}
\end{equation}

Our objective is to learn an effective representation $(\varPhi,  \varPsi)$ that can capture the underlying state transition dynamics in the latent space.
Specifically, the latent state-action embedding $z_{sa}$ can be interpreted as the predicted next latent state when taking action $a$ at state $s$. Let $s^{\prime}$ denote the corresponding next state in the original state space and $z_{s^{\prime}}$ its latent representation. We therefore expect that
\begin{equation}
	\begin{split}
z_{sa} = z_{s^{\prime}}, \forall s \in \mathcal{S}, a \in \mathcal{A}. 
\end{split}
\end{equation}

Therefore, the training objective of state and  state-action embedding is defined by minimizing the expected loss
\begin{equation}
	\begin{split}
     L(\varPhi,  \varPsi) & =\mathbb{E}_{s' \sim P(\cdot \mid s, a)} \left \Vert z_{sa} - z_{s^\prime}  \right\Vert \\
      & = \mathbb{E}_{s' \sim P(\cdot \mid s, a)} \left \Vert \varPsi(\varPhi(s), a) - z_{s^\prime}  \right\Vert
\end{split}
\end{equation}

\textbf{SR-enhanced DRL}: In the proposed SR-DRL, the inputs of actor and critic networks are augmented to involve the state and state-action embedding. Compared with conventional DRL, we have the following modifications
\begin{equation} 
	\begin{cases} 
		DRL     & \to \quad   SR-DRL \\
		\quad \pi(s) & \to \quad \pi\left(s, z_s\right) \\ 
		Q(s, a) & \to \quad  Q\left(s, a, z_s,  z_{sa}\right) 
	\end{cases} 
	\label{eq: encoded actor-critic} 
\end{equation}

The representation learning framework can be combined with any existing DRL methods that admit a actor-critic architecture, such as Deep Deterministic Policy Gradient (DDPG) \cite{lillicrap2015ddpg} and Twin Delayed Deep Deterministic Policy Gradient (TD3) \cite{fujimoto2018td3}.  TD3 introduces three improvements: double critic, delayed actor update, and target policy smoothing. In SR-DRL, representation learning and DRL training are decoupled and alternated. 
The gradients for the state and state-action embeddings are
\begin{equation}
	\begin{split}
		\nabla_{\varPhi} L(\varPhi, \varPsi)
		&= \mathbb{E}_{s' \sim P(\cdot \mid s, a)}
		\left[
		\nabla_{\varPhi}
		\left\|
		\varPsi(\varPhi(s), a) - \left\vert \varPhi(s') \right\vert_{\times}~
		\right\|
		\right] \\
		\nabla_{\varPsi} L(\varPhi, \varPsi)
		&= \mathbb{E}_{s' \sim P(\cdot \mid s, a)}
		\left[
		\nabla_{\varPsi}
		\left\|
		\varPsi(\varPhi(s), a) - \left\vert \varPhi(s') \right\vert_{\times}~
		\right\|
		\right]
	\end{split}
	\label{eq:embedding_loss}
\end{equation}
where we use $\vert \cdot \vert_{\times}$ to denote stop-gradient operation, which can  avoid  the collapse of representation learning during training.

\textbf{Actor and Critic Updates}: We denote the actor and target actor networks as $\pi^{\theta}, \pi^{\theta^{\prime}}$. The critic and target critic networks are  $Q^{\omega}, Q^{\omega^{\prime}}$. 
For DRL, the critic is trained to minimize the Bellman error, while the actor is updated via the deterministic policy gradient. The critic updates of DDPG and TD3 are presented as follows. 
\paragraph{DDPG Critic Updates:}
\begin{equation}
	\begin{cases}
	 L_{\rm critic}(\omega) = \mathbb{E}_{(s_t, a_t, r_t, s_{t+1}) \sim D}  
	\Big[ \big( Q^\omega(s_t, a_t, z_s, z_{sa}) - y_t \big)^2 \Big], \\
	 y_t = r_t + \gamma \, Q^{\omega'} \big( s_{t+1},  \pi^{\theta'}(s_{t+1}, z_{s'}), z_{s'}, z_{s'a'} \big).\\
	 \omega \gets \omega - \eta_Q \, \nabla_\omega L_{\rm critic}(\omega),
	\end{cases}
	\label{eq:ddpg_critic_update}
\end{equation}

\paragraph{TD3 Critic Updates:}  
\begin{equation}
	\begin{cases}
		L_{\rm critic}(\omega_i) = \mathbb{E} \Big[ \big( Q^{\omega_i}(s_t, a_t, z_s, z_{sa}) - y_t \big)^2 \Big], i = 1, 2 \\
	y_t \!=\! r_t \!+\! \gamma \, \min_{i=1,2} Q^{\omega'_i} \Big(s_{t+1}, \pi^{\theta'}(s_{t+1}, z_{s'}) \!+\! \epsilon, z_{s'}, z_{s'a'} \Big), \\
		\omega_i \gets \omega_i - \eta_Q \, \nabla_{\omega_i} L_{\rm critic}(\omega_i), \quad i=1,2
		\end{cases}
	\label{eq:td3_critic_update}
\end{equation}
where $\epsilon \sim \text{clip}(\mathcal{N}(0, \sigma), -c, c)$ is the clipped target policy noise, and $Q^{\omega'_1}, Q^{\omega'_2}$ are the two target critic networks.

\paragraph{DDPG/TD3 Actor Updates:}
\begin{equation}
	\begin{cases}
	\nabla_\theta J \approx \mathbb{E}_{s \sim D} \Big[ 
	\nabla_a Q^{\omega}(s_t, a_t, z_s, z_{sa}) \big|_{a = \pi^\theta(s, z_s)} \, 
	\nabla_\theta \pi^\theta(s, z_s)
	\Big], \quad  \\
	\theta \gets \theta + \eta_\pi \, \nabla_\theta J.
	\end{cases}
	\label{eq:actor_update}
\end{equation}

For target actor and critic networks, updates are often performed at a slower pace for stable training. The widely used soft update scheme is adopted in our framework. The overall algorithm for the proposed SR-DRL is summarized in Algorithm~\ref{alg:sr-drl}. Note that the above framework applies to all DRL methods adopting the same actor-critic architecture.

\begin{algorithm}[t]
\caption{Training Algorithm of the Proposed SR-DRL Framework}
\label{alg:sr-drl}
\begin{algorithmic}[1]
\STATE \textbf{Initialization:}
(1) Initialize actor network $\pi^{\theta}$, $\pi^{\theta'}$, critic $Q^{\omega}$, $Q^{\omega'}$, state embedding $\varPhi$, state-action embedding $\varPsi$;
(2) Initialize replay buffer $\mathcal{D}$.

\FOR{episode $=1$ to $N_{\rm episode}$}
    \STATE Reset environment and obtain initial state $s_0$; initialize random noise $\mathcal{N}_t$
    \FOR{time step $t=0$ to $T-1$}
        \STATE State embedding: $z_s = \varPhi(s_t)$
        \STATE Select action: $a_t = \pi^{\theta}(s_t, z_s) + \mathcal{N}_t$
        \STATE Execute $a_t$ and get $r_t$, $s_{t+1}$
        \STATE Store transition $(s_t, a_t, r_t, s_{t+1})$ in $\mathcal{D}$

        \IF{$|\mathcal{D}| > N_{\min}$}
            \STATE Sample a mini-batch $(s_t, a_t, r_t, s_{t+1}) \sim \mathcal{D}$

            \STATE \textbf{Update embedding} according to \eqref{eq:embedding_loss} 
            \STATE \textbf{Update critic} according to \eqref{eq:ddpg_critic_update} or \eqref{eq:td3_critic_update}
            \STATE \textbf{Update actor} according to \eqref{eq:actor_update}
            \STATE \textbf{Soft update target networks} $\theta', \omega'$
        \ENDIF
    \ENDFOR
\ENDFOR
\end{algorithmic}
\end{algorithm}

\section{Experiments}
\label{sec:Experiments and Results}
This section evaluates the proposed  model and enhanced DRL-based method for the optimal operation of an HMES based on  real-world datasets. 
The experimental configurations are summarized as follows.  The model parameters of the involved HESS (including a Siemens Silyzer 100 electrolyzer, a Swiss Hydrogen fuel cell, a compressor, and a pressurized hydrogen storage tank)  are adopted from the experimental studies \cite{flamm2021electrolyzer, fochesato2024peak}.   
The configurations of other energy generation, conversion and storage devices follow the specifications of  \cite{dong2022optimal, liu2021optimal}. 
The time-of-use (ToU) electricity price and multi-energy demand profiles (electricity, heating, and cooling) of a building community are taken from the CityLearn dataset~\cite{nweye2025citylearn}. 
We assume symmetric purchasing and selling prices of electricity market.  
The hydrogen market is not explicitly considered in the studies but can be readily incorporated into the framework. The penalty factor for constraints are determined  by experience and set as $0.1$. For ease of reference, all system parameters are summarized in Table~\ref{tab:system_configuration}.  The following  methods for the optimal operation of HMES are compared in the experiments: 
\begin{itemize}
	\item \texttt{Perfect MPC}: 
	A deterministic model predictive control (MPC) approach is implemented by solving the optimization problem 
	$\{\min J_{\rm cost} ~{\rm s.t.}~ \eqref{eq:ESS} - \eqref{eq:energy_balance}\}$
	with  deterministic renewable supply and multi-energy demand profiles, representing the theoretical optima.
	
	\item \texttt{DDPG} and \texttt{TD3}: 
	Standard DDPG and TD3 agents are trained based on the formulated MDP to enable learning-based optimal operation of  HMES.
	
	\item \texttt{SR-DDPG} and \texttt{SR-TD3}: 
	The proposed state-action representation learning (SR) module is integrated with DDPG and TD3, and trained for the optimal operation of HMES.
\end{itemize}

For model and method evaluation, three representative cases of different time periods (i.e., \texttt{Case~1}, \texttt{Case~2}, and \texttt{Case~3}) are considered. For each case, a total of 30 days of data are used for training and evaluation. For the DRL (\texttt{DDPG} and \texttt{TD3}) and SR-DRL (\texttt{SR-DDPG} and \texttt{SR-TD3}), 20 days are used for training and  10 days for testing. The model architecture and training hyperparameters are summarized in Table~\ref{tab:hyperparameters}. For \texttt{Perfect MPC}, the optimal control policy for each day is obtained using the Gurobi solver. 

\begin{figure*}[htbp]
	\centering
	\includegraphics[width=0.95\textwidth, height=0.18\textheight, keepaspectratio]{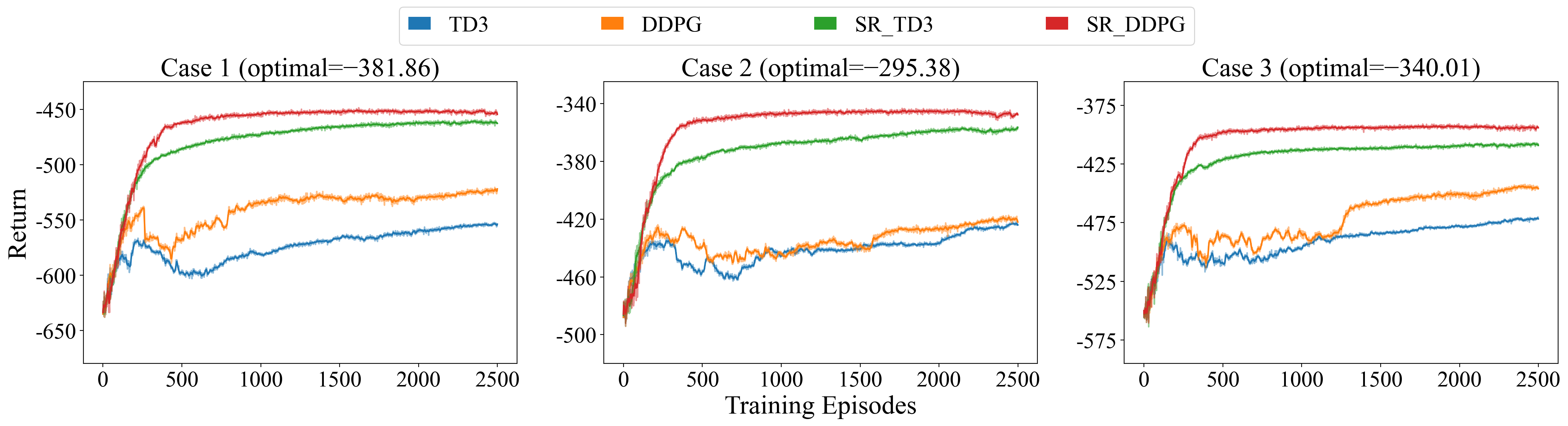}
	\captionsetup{skip=-2pt}
	\caption{The evolution of episode return during training under different methods.}
	\label{fig:training_convergence}
\end{figure*}

\subsection{Training performance}
\label{sec:RL training}

This section investigates the training performance of the DRL (\texttt{DDPG} and  \texttt{TD3}) and SR-DRL (\texttt{SR-DDPG} and  \texttt{SR-TD3}) methods. For fair comparisons, all methods share identical model and training parameters except for the state-action representation modules used by the SR-DRL methods.  For all cases,  the evolution of episode return during training are recorded and visualized in Figure~\ref{fig:training_convergence}.  The theoretical optima for each case provided by \texttt{Perfect MPC} by averaging the daily cost  are indicated in the figure titles. 
The results show that the SR-DRL methods consistently outperform conventional DRL in terms of both convergence rate and final episode return across all cases.  Moreover, SR-DRL approaches exhibit a markedly more stable learning process compared with their conventional counterparts.  
Specifically,  the SR-DRL methods (\texttt{SR-TD3} and \texttt{SR-DDPG}) improve final episode  returns  by 16.7\% and 13.4\% (\texttt{Case 1}),   15.5\% and 17.2\% (\texttt{Case 2}), 13.5\% and 11.5\% ( \texttt{Case 3}) compared with their conventional counterparts (\texttt{TD3} and \texttt{DDPG}). 
Notably,  these results highlight the superior capability of  SR-DRL approaches in  handling the complexity of engineering systems. For conventional \texttt{TD3}, the performance gaps relative to the theoretical optima are 45.1\% (\texttt{Case 1}), 43.3\% (\texttt{Case 2}), and 38.7\% (\texttt{Case 3}). In contrast, the proposed \texttt{SR-TD3} significantly reduces these gaps to 20.9\%, 21.1\%, and 20.1\%, respectively. Similar performance improvements are observed with the DDPG methods. While  conventional \texttt{DDPG} exhibits performance gaps of 37.1\%, 42.3\%, and 31.0\% across the three cases, the proposed \texttt{SR-DDPG} reduces these gaps to 18.7\%, 17.8\%, and 15.9\%, respectively. These results demonstrate the superior capability of SR-DRL methods in approaching high-quality control policies for complex networked energy systems.

\subsection{Operating performance}

We test the performance of trained DRL and SR-DRL agents for the optimal operation of HMES using the 10-day testing datasets. 
The average daily episode return, operation cost, and penalty incurred by  constraint violations are used as performance metrics  and  indicated by  \texttt{Cost +  $\lambda \cdot$ Penalty}, \texttt{Cost}, and \texttt{Penalty}, respectively. The results for all three cases with the considered performance metrics under the different methods  are reported    in Table~\ref{tab:test_comparison} with the results with our proposed methods highlighted in bold.  
The results show that the  SR-DRL methods (\texttt{SR-TD3} and \texttt{SR-DDPG}) apparently outperform  conventional DRL (\texttt{TD3} and \texttt{DDPG}) considering all performance  metrics.  
Specifically, for \texttt{Case 1}, the overall performance gaps  of DRL, measured by  \texttt{Cost +  $\lambda \cdot$ Penalty},   are 42.8\% (\texttt{TD3}) and 36.5\% (\texttt{DDPG}), which are reduced to 25.4\% and 21.0\% by the SR-DRL methods (\texttt{SR-TD3} and \texttt{SR-DDPG}). 
Similarly, the operation cost gaps are reduced from 39.1\% (\texttt{TD3}) and 32.1\% (\texttt{DDPG}) to 24.0\% (\texttt{SR-TD3}) and 20.3\% (\texttt{SR-DDPG}), respective.
 Similar improvements can be  observed for \texttt{Case 2} and \texttt{Case 3}.
Notably, the SR-DRL methods exhibit superior advantages in handling system constraints  as substantial reduction of penalties incurred by constraint violations are observed with the SR-DRL methods compared with the DRL counterparts across all cases.

\begin{figure}[htbp]
	\centering
	\includegraphics[width=0.95\linewidth, height=0.3\textheight]{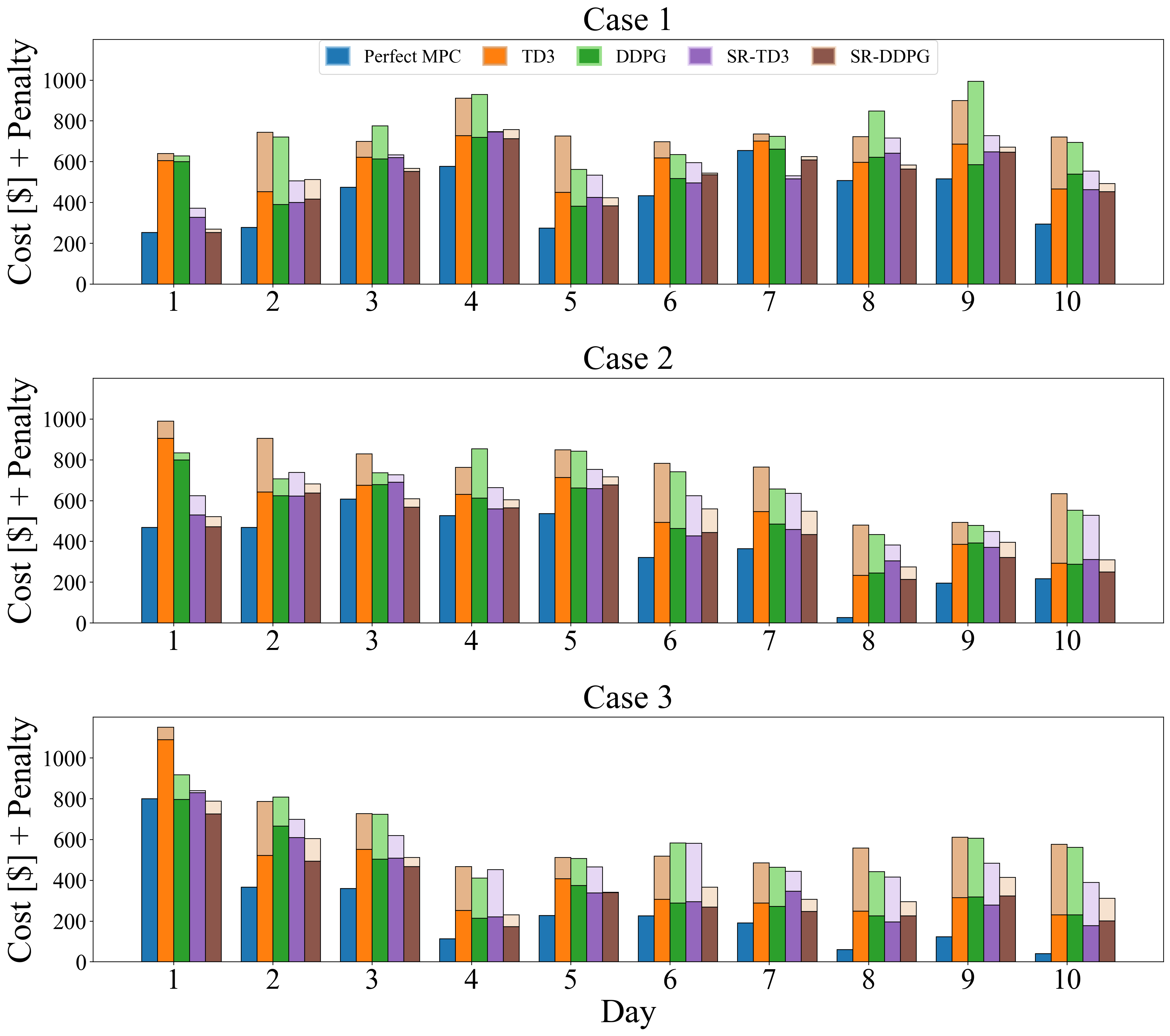}
	\captionsetup{skip=-1pt}
	\caption{Distribution of stacked cost and penalty of constraint violations under different methods.}
	\label{fig:distribution}
\end{figure}

The superior performance of the SR-DRL methods over conventional DRL is further evidenced by the distributions of operational cost and constraint violation penalties shown in Figure~\ref{fig:distribution}.
Since the \texttt{Perfect MPC} assumes complete system information and explicitly enforces all constraints, no constraint violation penalties are observed across all cases.
In contrast, both DRL and SR-DRL methods incur certain penalties due to system uncertainties. Notably, the SR-DRL methods (\texttt{SR-TD3} and \texttt{SR-DDPG}) consistently achieve lower daily operating costs and reduced constraint violation penalties compared with their DRL counterparts (\texttt{TD3} and \texttt{DDPG}).

\begin{table}[htbp]
\centering
\small
\caption{Performance of DRL and SR-DRL methods compared with Perfect MPC for different cases}
	\setlength{\tabcolsep}{1.8pt}
\begin{tabular}{llccccc}
\toprule
\multirow{2}{*}{Cases} & \multirow{2}{*}{Methods} & \multicolumn{2}{c}{\small Cost + $\lambda \cdot$ Penalty} & \multicolumn{2}{c}{\small Cost} & \multirow{2}{*}{\small Penalty} \\ \cmidrule(lr){3-4} \cmidrule(lr){5-6}
 &  & [\$] & Gap[\%] & [\$] & Gap[\%] &  \\ \midrule

% Case 1
\multirow{5}{*}{Case 1} & Perfect MPC & 426.44 & -- & 426.44 & -- & 0 \\
 \cline{2-7}
 & TD3 & 608.83 & 42.8 & 593.15 & 39.1 & 156.77 \\
 & \textbf{SR-TD3} & \textbf{534.92} & \textbf{25.4} & \textbf{528.62} & \textbf{24.0} & \textbf{63.01} \\
 \cline{2-7}
 & DDPG & 582.11 & 36.5 & 563.30 & 32.1 & 188.14 \\
 & \textbf{SR-DDPG} & \textbf{516.17} & \textbf{21.0} & \textbf{512.97} & \textbf{20.3} & \textbf{31.96} \\ \midrule

% Case 2 
\multirow{5}{*}{Case 2} & Perfect MPC & 373.34 & -- & 373.34 & -- & 0 \\
 \cline{2-7}
 & TD3 & 572.04 & 53.2 & 552.35 & 47.9 & 196.88 \\
 & \textbf{SR-TD3} & \textbf{505.48} & \textbf{35.4} & \textbf{493.59} & \textbf{32.2} & \textbf{118.90} \\
 \cline{2-7}
 & DDPG & 541.31 & 45.0 & 525.48 & 40.8 & 158.26 \\
 & \textbf{SR-DDPG} & \textbf{464.88} & \textbf{24.5} & \textbf{458.49} & \textbf{22.8} & \textbf{63.90} \\ \midrule

% Case 3
\multirow{5}{*}{Case 3} & Perfect MPC & 250.57 & -- & 250.57 & -- & 0 \\
 \cline{2-7}
 & TD3 & 443.36 & 76.9 & 421.56 & 68.2 & 217.97 \\
 & \textbf{SR-TD3} & \textbf{395.97} & \textbf{58.0} & \textbf{380.04} & \textbf{51.7} & \textbf{159.27} \\
  \cline{2-7}
 & DDPG & 410.61 & 63.9 & 389.23 & 55.3 & 213.78 \\
 & \textbf{SR-DDPG} & \textbf{353.75} & \textbf{41.2} & \textbf{346.68} & \textbf{38.4} & \textbf{70.70} \\ \bottomrule
\end{tabular}
\label{tab:test_comparison}
\end{table}

\begin{figure*}[htbp]
   \centering
   \begin{subfigure}{0.33\linewidth}
       \centering
       \includegraphics[width=\linewidth, height=0.28\textheight]{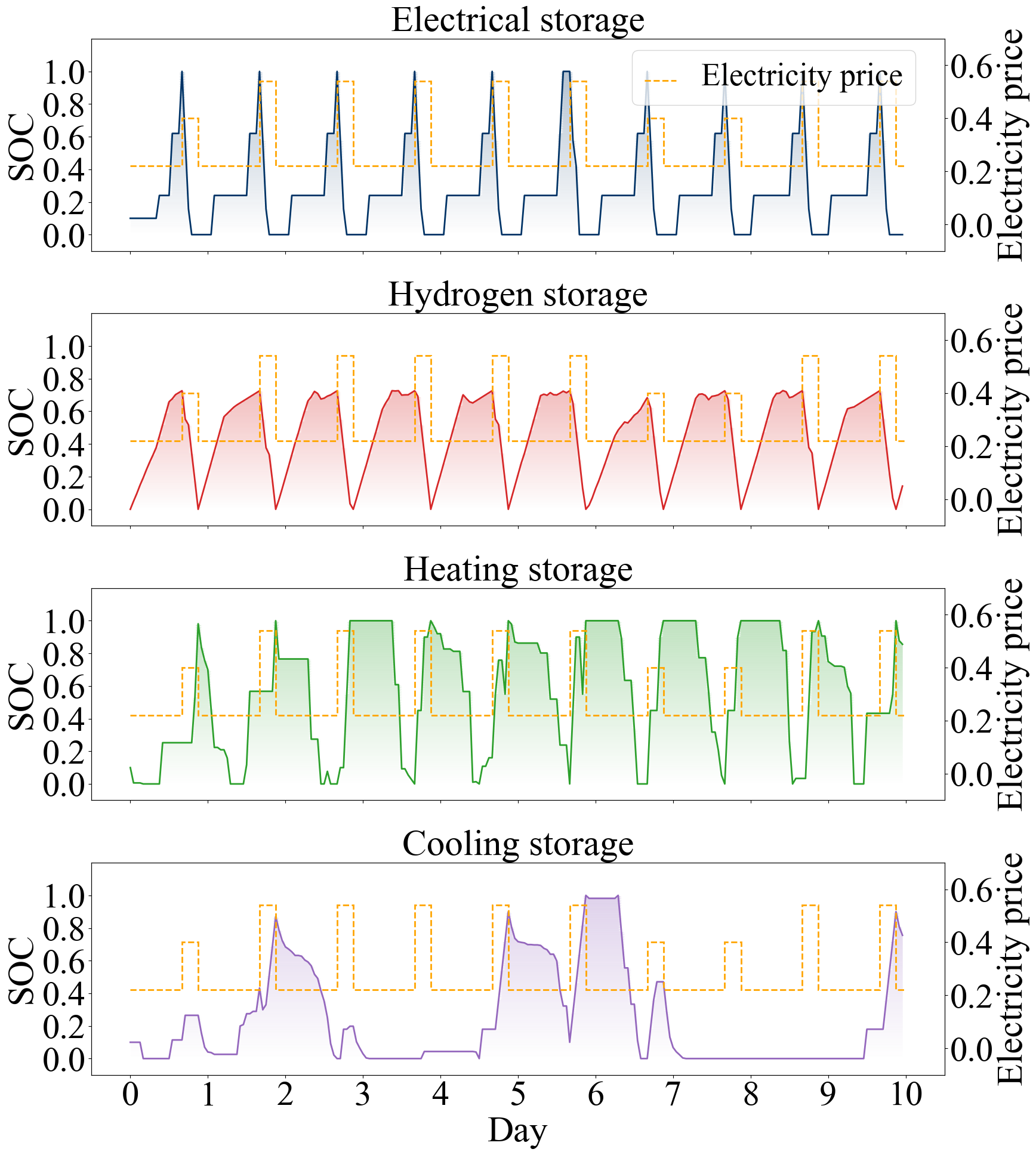}
	   \captionsetup{skip=1pt}
       \caption{Perfect MPC}
       \label{fig:strategies_optimal_2}
   \end{subfigure}
   \hfill
   \begin{subfigure}{0.33\linewidth}
       \centering
       \includegraphics[width=\linewidth, height=0.28\textheight]{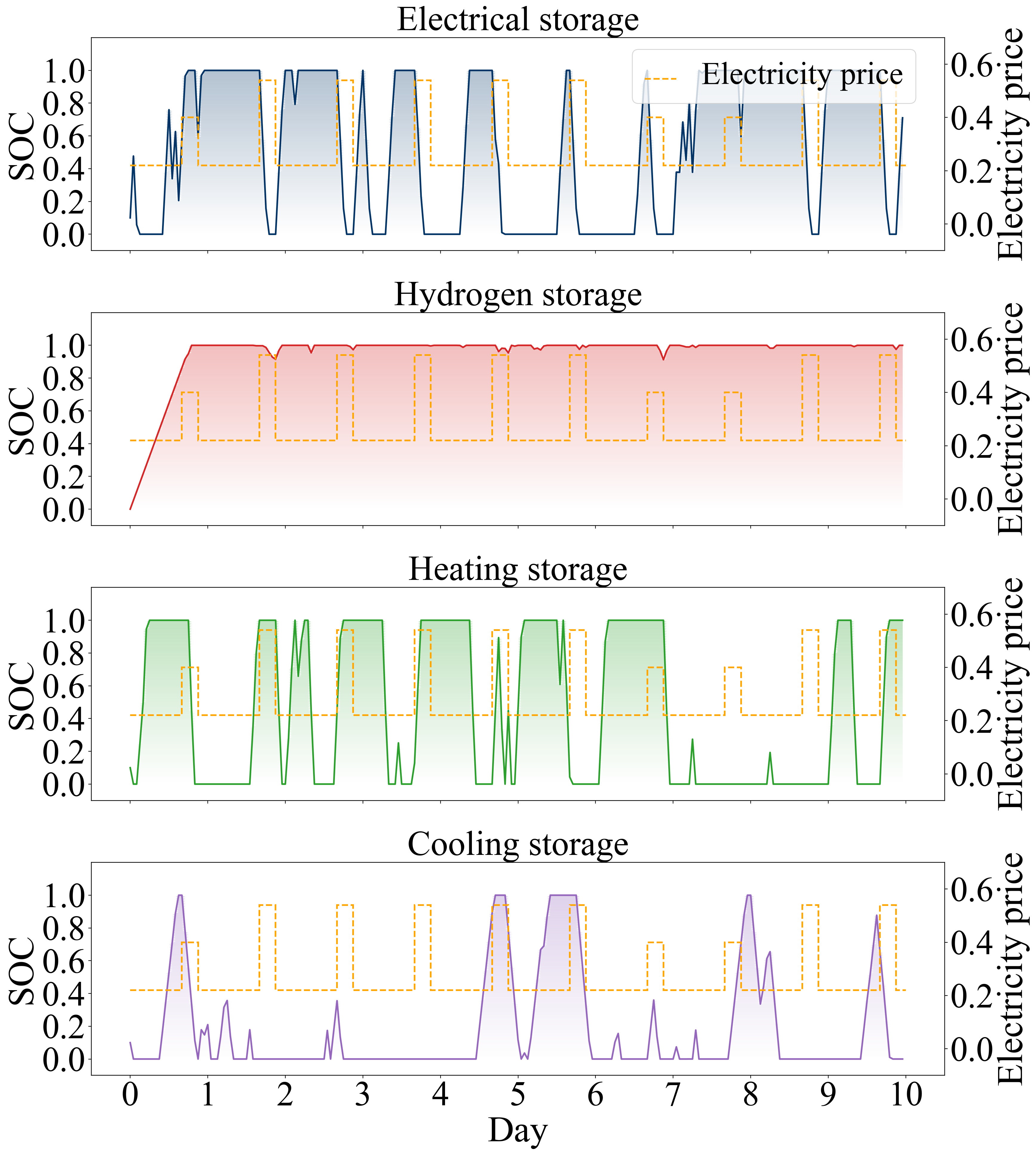}
	   \captionsetup{skip=1pt}
       \caption{TD3}
       \label{fig:strategies_td3}
   \end{subfigure}
   \hfill
   \begin{subfigure}{0.33\linewidth}
       \centering
       \includegraphics[width=\linewidth, height=0.28\textheight]{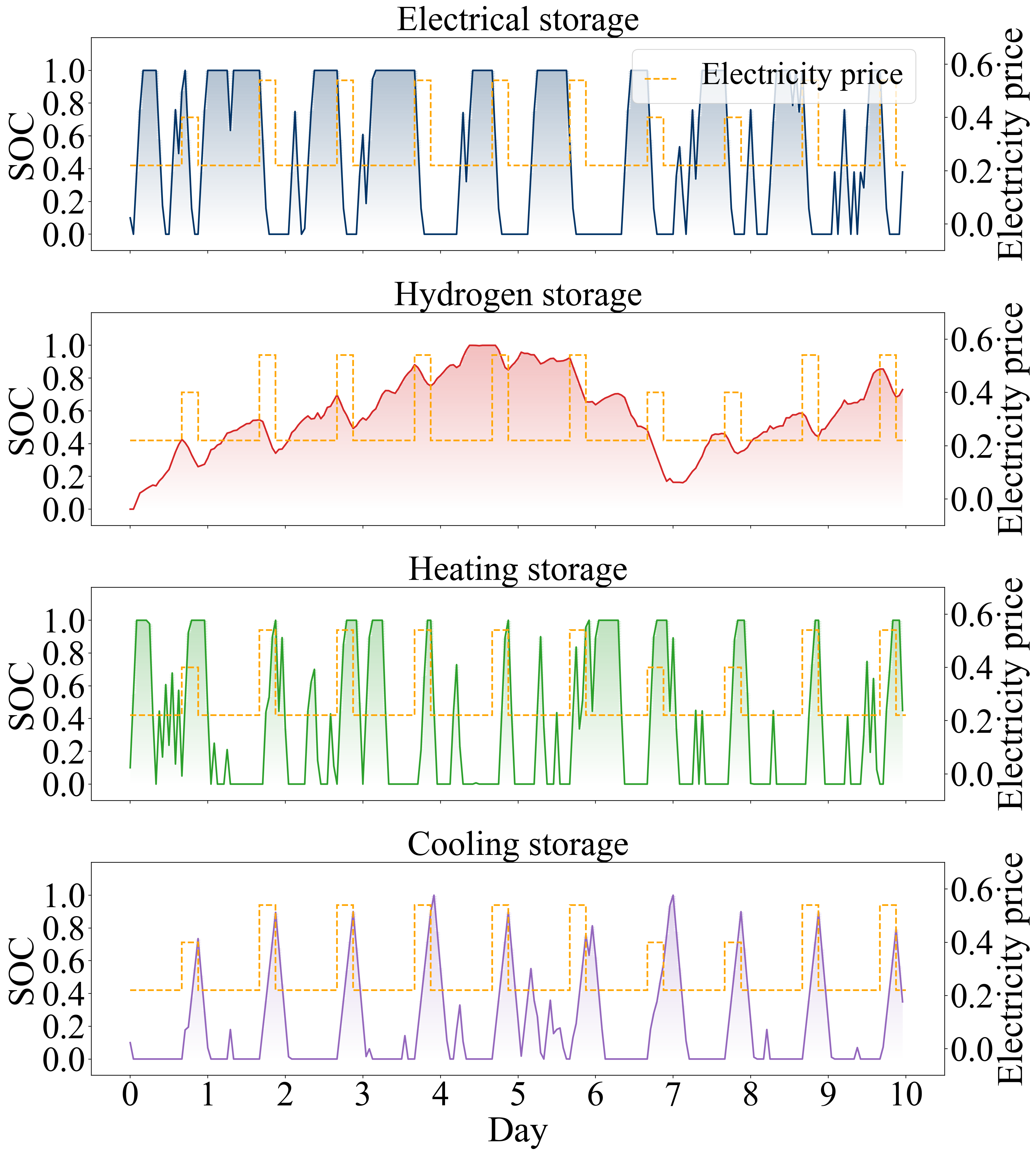}
	   \captionsetup{skip=1pt}
       \caption{SR-TD3}
       \label{fig:strategies_sr_td3}
   \end{subfigure}
   \captionsetup{skip=-2pt}
   \caption{The evolution of SOC of different types of energy storage devices in response to electricity market price with different methods.}
   \label{fig:policies_td3}
\end{figure*}

To further investigate the operational performance of different methods, we examine the responses of all energy storage devices (i.e., ESS, TES, CES, and HSS) to the time-varying electricity prices. We use \texttt{Case 1} and the TD3 methods as an example and present the evolution of state-of-charge (SOC) of all energy storage devices in Figure \ref{fig:policies_td3}. Overall, all storage devices exhibit clear price-sensitive behaviors: they generally charge during periods of low electricity prices and discharge when prices are high, which is consistent with our operating cost reduction objective. 
Whereas, for the DRL methods (\texttt{TD3}),  the hydrogen storage does not respond well  compared with the proposed SR-DRL (\texttt{SR-TD3}). Specifically, the hydrogen storage remains close to full over the testing days and  is not effectively discharged during high-price periods, when storage backup is expected to function for operating cost reduction. 
This actually demonstrates  the limitations of conventional DRL methods for  handling complex engineering systems with strong spatial-temporal couplings. 

\subsection{Model evaluation}
\begin{figure}[htbp]
    \centering
    % ================= Row 1 =================
    \begin{subfigure}{0.95\linewidth}
        \centering
        \includegraphics[width=\linewidth, height=0.12\textheight]{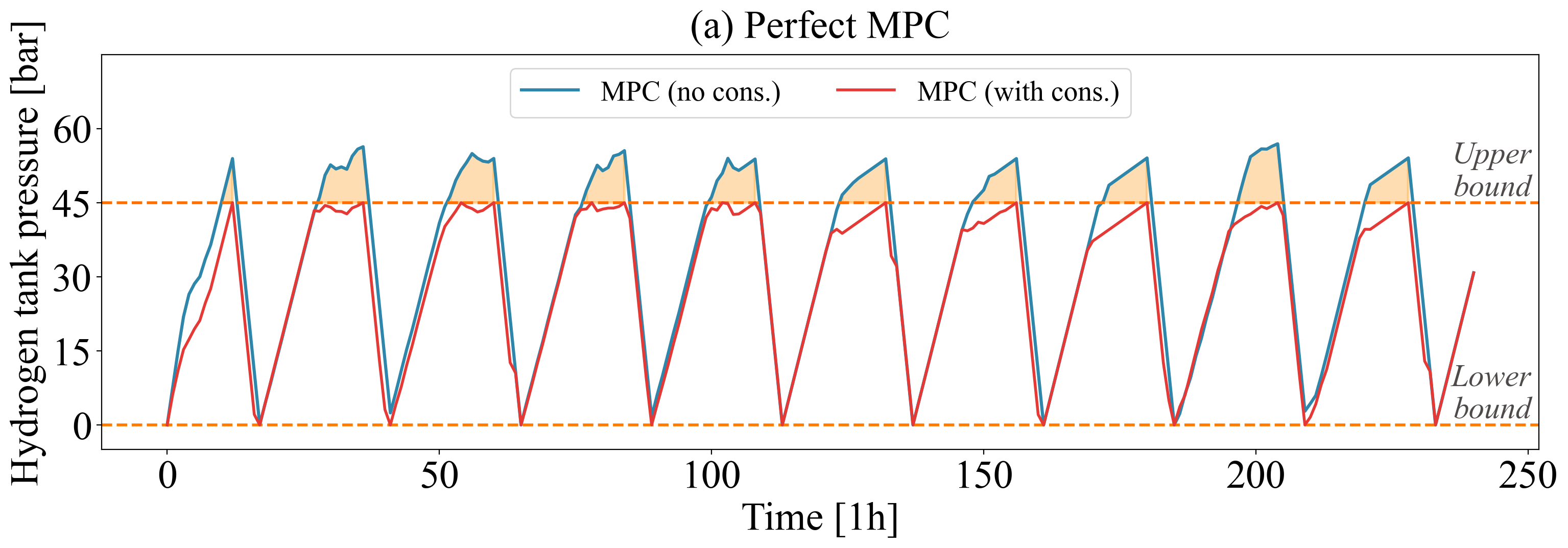}
        % \caption{Operational behaviors for perfect MPC}
        % \label{subfig:mpc_p_tank}
    \end{subfigure}
    \begin{subfigure}{0.95\linewidth}
        \centering
        \includegraphics[width=\linewidth, height=0.12\textheight]{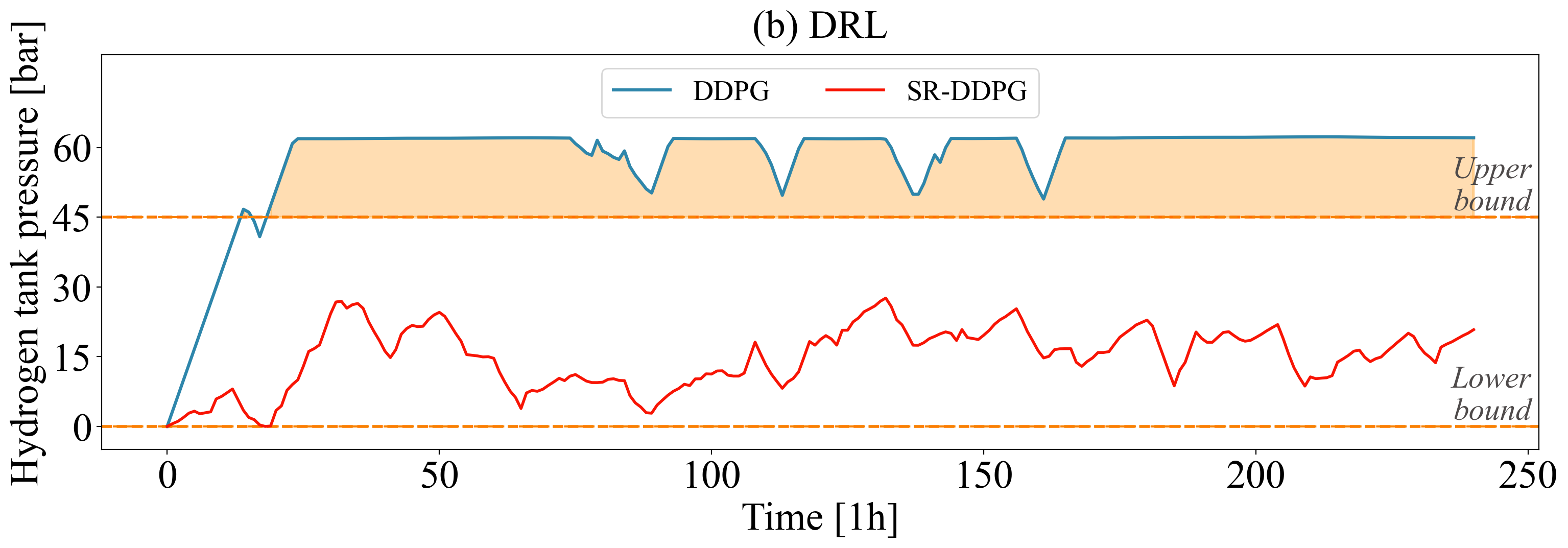}
        % \caption{Operational behaviors for DDPG and SR-DDPG}
        % \label{subfig:rl_p_tank}
    \end{subfigure}
	\caption{Evolution of hydrogen tank pressure over the 10-day testing periods.}
	\label{fig:mpc_drl_compare}
\end{figure}

This section evaluates the importance of explicitly considering the multi-physics coupled processes of the HESS for ensuring the safe and reliable operation of the HMES. For the proposed model, the safety-critical  variables of the HESS, including  stack current, stack temperature, and hydrogen tank pressure  are explicitly modeled.  
We  first examine the proposed model with \texttt{Perfect MPC} by comparing two scenarios: with and without the safety constraints on stack temperature, stack current, and tank pressure, as defined in \eqref{eq:EL_T}, \eqref{con:EL_current}, and \eqref{con:tank_pressure_limits} with the 10-day testing datasets of \texttt{Case 2}. Omitting these constraints is equivalent to neglecting the multi-physics interactions of current, temperature, and pressure with the HESS.  For the considered case, we note that 
though  the stack temperature and current mostly remain within their feasible ranges with both scenarios, the hydrogen tank pressure exhibits severe violations when the safety constraints are removed as shown in Figure~\ref{fig:mpc_drl_compare}(a).
Whereas the tank pressure has been strictly maintained within the allowable limits when the safety constraints are explicitly enforced. This demonstrates the importance of explicitly considering  the  multi-physics process of HESS to ensure  reliable operation of the HMES.  

We further evaluate the effectiveness of the proposed SR-DRL methods in ensuring the safe operation of HESS under uncertainties. Specifically, we examine the evolution of hydrogen tank pressure over a 10-day testing period using \texttt{DDPG} and \texttt{SR-DDPG}. As shown in Figure~\ref{fig:mpc_drl_compare}(b), conventional \texttt{DDPG} frequently violates the pressure limits, whereas \texttt{SR-DDPG} consistently maintains safe operation throughout the period. These results highlight the advantage of SR-DRL in enforcing physical constraints, which is essential for the practical deployment of learning-based control strategies in HMES.

\subsection{Function of Representation Learning}
In this section, we provide some insights on the role of state-action representation learning (SR) for DRL. we rely on principal component analysis (PCA) and clustering analysis for intuitive illustrations. The functions of SR can be interpreted from two perspectives.

\begin{figure}[htbp]
	\centering
	\includegraphics[width=0.48\textwidth, height=0.39\textheight]{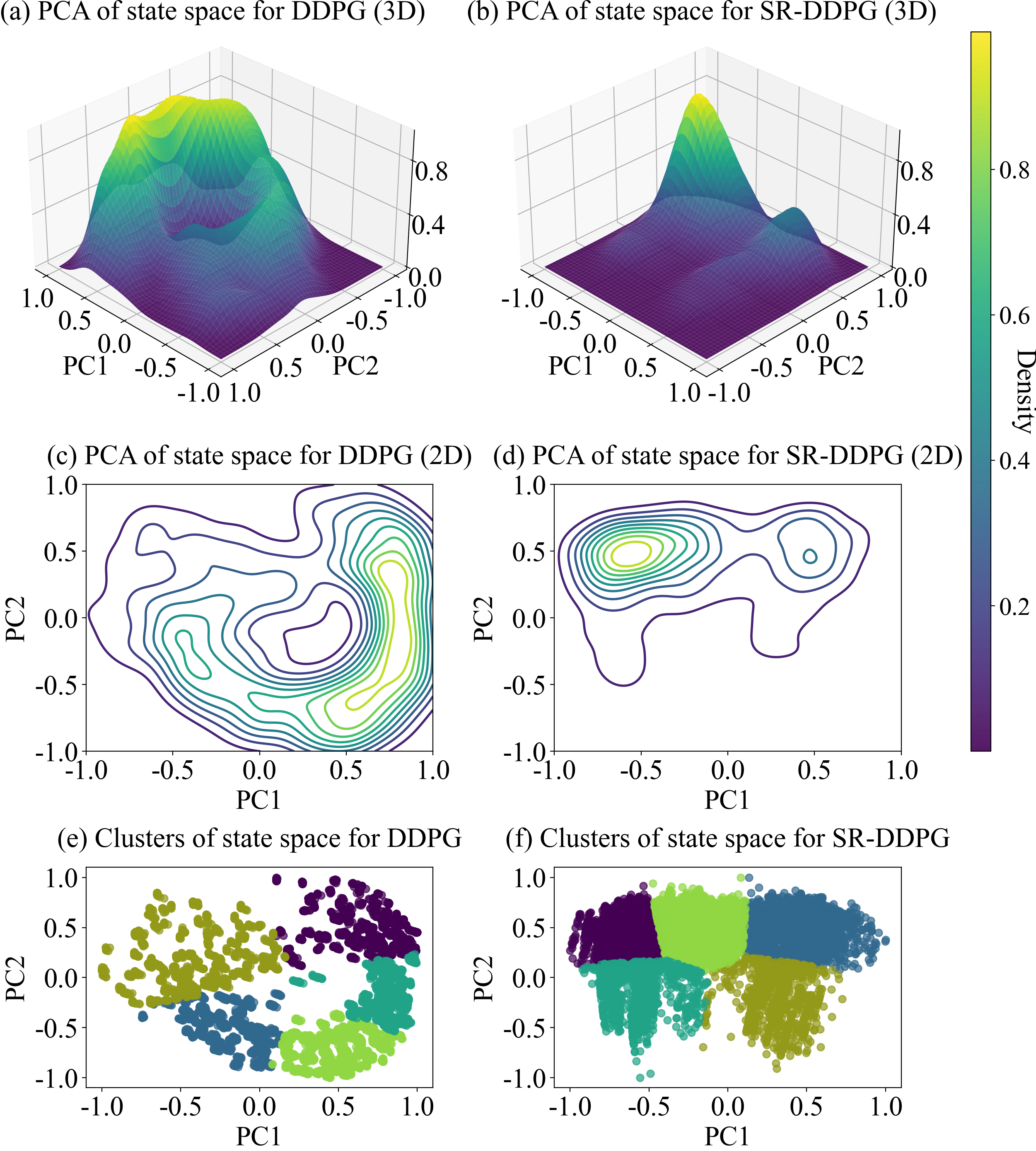}
	\caption{3D, 2D, and clustering analysis of PCA-projected state space with DDPG and SR-DDPG.}
	\label{fig:interpret}
\end{figure}

\emph{First, representation learning can reshape the geometric structure of state space, making it more amenable to policy and value function learning in DRL.} To illustrate this effect, we analyze the state-space distributions of DRL (\texttt{DDPG}) and SR-DRL (\texttt{SR-DDPG}) using PCA. We compare the outputs of the first layer of the actor networks of DRL and SR-DRL, which respectively convert the raw states $s$ and augmented state $[s, z_s]$ into the same dimensionality and enable a fair comparison. A total of 10,000 state samples are collected and projected onto the first two principal components, with the resulting 3D and 2D visualizations shown in Figure~\ref{fig:interpret}(a)-(d).
For conventional DDPG, we observe that the state space exhibits a multi-scattered and irregular geometric structure, characterized by multiple  high-density regions. Such a poorly structured state space  is expected to pose challenges for the actor-critic learning of DRL, as it hinders discriminative feature extraction and increases the difficulty of accurately approximating the policy and value functions, ultimately slowing convergence.
In contrast, when the state-action representation learning modules are incorporated with SR-DDPG, the resulting state space exhibits a compact and well-structured geometry, which likely underlies the improved convergence and overall performance observed with SR-DRL approaches. 

\emph{Second, representation learning seems capable of reorganizing the state space from a decision-oriented perspective.} To examine this effect, we further conduct a clustering analysis on the PCA-projected state space with the DRL and SR-DRL methods. For a fair comparison, the state spaces are normalized and the resulting clusters are displayed  
in Figure~\ref{fig:interpret}(e) and (f).  We note that, for conventional DRL,  the clusters of state space are quite loose and diffuse, indicating the presence of substantial task-irrelevant noise. 
In contrast, the SR-DRL method reorganizes the states  into compact and well-separated clusters with clear boundaries. Each cluster corresponds to a coherent decision-relevant mode, within which states share similar physical interpretations and control implications. 
From a learning perspective, such a cluster-aware state space is expected to significantly reduces the complexity of policy and value function learning, improves learning stability and convergence rate.

\section{Conclusion and future work}
\label{sec:conclusion}
This paper investigates the optimal operation of a hydrogen-based multi-energy system (HMES) that integrates diverse energy generation, conversion, and storage devices to provide coordinated electricity, heating, and cooling services for a building community. We propose a comprehensive operational model that explicitly captures the nonlinear operating characteristics and multi-physics processes of hydrogen energy storage system (HESS), which are crucial for safe and reliable HMES operation but have been mostly overlooked.

To address computational challenges posed by complex system dynamics, strong spatial-temporal couplings, and multiple uncertainties, a representation-learning enhanced deep reinforcement learning (SR-DRL) approach is proposed, significantly accelerating convergence rates and improving performance over conventional DRL. We further give some illustrative insights on the role of representation learning for DRL based on principal component analysis and clustering analysis of a state space, concluding that it can reshape the state space of HMES into a well-structured and clustered geometry, facilitating the actor-critic learning with DRL methods. 
 
Future promising directions include the extension of the representation learning approach to enhance multi-agent deep reinforcement learning with its application to the coordinated operation of interconnected energy systems.

% \section{Sample Appendix Section}
% \label{sec:sample:appendix}
% Lorem ipsum dolor sit amet, consectetur adipiscing elit, sed do eiusmod tempor section \ref{sec:sample1} incididunt ut labore et dolore magna aliqua. Ut enim ad minim veniam, quis nostrud exercitation ullamco laboris nisi ut aliquip ex ea commodo consequat. Duis aute irure dolor in reprehenderit in voluptate velit esse cillum dolore eu fugiat nulla pariatur. Excepteur sint occaecat cupidatat non proident, sunt in culpa qui officia deserunt mollit anim id est laborum.

\bibliographystyle{elsarticle-num} 
\bibliography{cas-refs}

\clearpage
\appendix 
\section{Parameters}

\noindent
\begin{minipage}{\textwidth}
% \begin{table*}[!h]
	\centering
	\captionof{table}{Energy system configuration and device parameters}
	\label{tab:system_configuration}
	\setlength{\tabcolsep}{3pt}
	\renewcommand{\arraystretch}{0.95}
	\begin{tabular}{p{2cm} p{2.5cm} p{2cm} p{2cm} p{2.5cm} p{2cm} p{2cm}}
		\toprule
		Device & Param. & Value & Units & Param. & Value & Units \\
		\hline
		\multirow{3}{*}{ESS} & 
		$S^{\rm ess, \min}$ & 0 & kWh & $S^{\rm ess, \max}$ & 50 & kWh \\
		& $P^{\rm ess,ch}_{\max}$ & 20 & kW &  $P^{\rm ess,dis}_{\max}$ & 20 & kW \\
		& $S^{\rm ess}_{\rm init}$ & 5 & kWh & $\eta^{\rm ess, ch/dis}$ & 0.95 & -- \\
		\hline
		\multirow{3}{*}{TES} & 
		$S^{\rm tes, \min}$ & 0 & kWh & $S^{\rm tes, \max}$ & 100 & kWh \\
		& $S^{\rm tes}_{\rm init}$ & 10 & kWh & $g^{\rm tes,ch}_{\max}$ & 50 & kW \\
		& $g^{\rm tes,dis}_{\max}$ & 50 & kW & $\eta^{\rm tes, ch/dis}$ & 0.9 & -- \\
		\hline
		%		\multicolumn{6}{c}{\textbf{Cooling Energy System}} \\
		%		\hline
		\multirow{3}{*}{CES} & 
		$S^{\rm ces, \min}$ & 20 & kWh & $S^{\rm ces, \max}$ & 200 & kWh \\
		& $S^{\rm ces}_{\rm init}$ & 20 & kWh & $q^{\rm ces,ch}_{\max}$ & 40 & kW \\
		& $q^{\rm ces,dis}_{\max}$ & 40 & kW & $\eta^{\rm ces, ch/dis}$ & 0.9 & -- \\
		\hline
		\multirow{10}{*}{EL} & 
		$P^{\rm ely}_{\min}$ & 15 & kW & $P^{\rm ely}_{\max}$ & 200 & kW \\
		& $P^{\rm ely}_{\rm nom}$ & 100 & kW & $P^{\rm ely}_{\rm nom, 1}$ & 25 & kW \\
		& $z_1$ & $1.618 \cdot 10^{-5}$ & ${\rm m^3}/{[^\circ{\rm C} \cdot s]}$ & $z_0$ & $1.490 \cdot 10^{-2}$ & $\rm {m^3}/{s}$ \\
		& $z_{\rm low}$ & $1.530 \cdot 10^{-4}$ & $\rm m^3 / [\rm s \cdot kW]$ & $z_{\rm high}$ & $1.195 \cdot 10^{-4}$ & $\rm m^3 / [\rm s \cdot kW]$ \\
		& $T^{\rm ely}_{\max}$ & 70 & $^\circ$C & $T^{\rm ely}_{\rm init}$ & 20 & $^\circ$C \\
		& $j_0$ & 3.958 & $^\circ$C & $j_1$ & 0.551 & -- \\
		& $j_2$ & 0.430 & $\rm ^\circ C / kWh$ & $h_0$ & 235.254 & $^\circ$C \\
		& $h_1$ & 0.673 & -- & $h_{\rm low}$ & 0.987 & $\rm A / kW$ \\
		& $h_{\rm high}$ & 9.0 & $\rm A / kW$ & $i^{\rm ely}_{\rm nom}$ & 300 & A \\
		& $C^{\rm ely}_{\max}$ & 75 & Ah & $\alpha$ & 0.697 & -- \\
		\hline
		\multirow{7}{*}{HT} & 
		$S^{\rm hss, \min}$ & 0 & kg & $S^{\rm hss, \max}$ & 50 & kg \\
		& $S^{\rm hss}_{\rm init}$ & 0 & kg & $V^{\rm tank}$ & 10 & m$^3$ \\
		& $p^{\rm tank}_{\min}$ & 0 & bar & $p^{\rm tank}_{\max}$ & 45 & bar \\
		& $m^{\rm hss,ch}_{\max}$ & 30 & $\rm m^3 / h$ & $m^{\rm hss,dis}_{\max}$ & 30 & $\rm {m^3}/{h}$ \\
		& $\eta^{\rm hss, ch/dis}$ & 1.0 & -- & $\rho_0$ & 8.99$\cdot 10^{-2}$ & $\rm kg / m^3$ \\
		& $b_0$ & 11.5$\cdot 10^5$ & $\rm {m^2}/{s^2}$ & $b_1$ & 4.16$\cdot 10^3$ & $\rm {m^2}/[^\circ C\cdot s^2]$ \\
		& $g_0$ & 0.94 & -- & $g_1$ & 5.91$\cdot 10^{-2}$ & -- \\
		\hline
		\multirow{5}{*}{FC} & 
		$P^{\rm fc}_{\max}$ & 160 & kW & $P^{\rm fc}_{\min}$ & 0 & kW \\
		& $P^{\rm fc}_{\rm bp}$ & 47.97 & kW & $\eta_{\rm fc}$ & 0.3 & -- \\
		& $\eta^{\rm fc}_{\rm rec}$ & 0.8 & -- & $i^{\rm fc}_{\rm bp}$ & 122.80 & A \\
		& $s_1$ & 2.56 & $\rm 1 / kV$ & $s_2$ & 3.31 & $\rm 1 / kV$ \\
		& $c$ & 0.21 & $\rm {Nm^3}/{C}$ & -- & -- & -- \\
		\hline
		%		\multicolumn{6}{c}{\textbf{Absorption Chiller}} \\
		%		\hline
		\multirow{1}{*}{AC} & 
		$g^{\rm ac}_{\max}$ & 200 & kW & $\eta_{\rm ac}$ & 0.94 & -- \\
		\hline
		%		\multicolumn{6}{c}{\textbf{Renewable Energy and Markets}} \\
		%		\hline
		\multirow{3}{*}{PV} & 
		$A_{\rm PV}$ & 1500 & m$^2$ & $A_{\rm stc}$ & 400 & m$^2$ \\
		& $\eta^{\rm stc}$ & 0.762 & -- & $\eta^{\rm PV}$ & 0.2 & -- \\
		\bottomrule
	\end{tabular}

	\vspace{2em}

	\captionof{table}{DRL and SR-DRL model and training parameters}
	\label{tab:hyperparameters}
	\renewcommand{\arraystretch}{0.95} % 调整行高
	\setlength{\tabcolsep}{5pt}
	\begin{tabular}{l@{\hspace{2em}}l@{\hspace{2.5em}}l@{\hspace{2em}}l@{\hspace{2.5em}}}
	\toprule
	Parameter & Value & Parameter & Value \\
	\midrule
	Optimizer & Adam & Epochs & 2500 \\
	Batch size & 1024 & Buffer size & $10^6$ \\
	Discount rate & 1.0 & Learning rate & $3\times 10^{-4}$ \\
	Soft update rate & 0.01 &  \\
	State embedding & [256, 256, 256, 512] & State-action embedding & [256, 256, 256, 512] \\
	Actor & [512, $\rm action\_dim$] &
	Critic & [512, 512, 1] \\
	\bottomrule
	\end{tabular}
\end{minipage}

% \begin{table*}[ht]
% 	\centering
% 	\caption{DRL and SR-DRL model and training parameters}
% 	\label{tab:hyperparameters}
% 	\renewcommand{\arraystretch}{0.95}
% 	\begin{tabular}{l@{\hspace{3em}}l}
% 		\toprule
% 		Parameter & Value \\
% 		\midrule
% 		Optimizer & Adam \\
% 		Epochs & 2500 \\
% 		Batch size & 1024 \\
% 		Buffer size & $10^6$ \\
% 		Discount rate & 1.0 \\
% 		Learning rate & $3\times 10^{-4}$ \\
% 		Soft update rate  & 0.01 \\
% 		\midrule
% 		State embedding & [256, 256, 256, 512]\\
% 		State-action embedding & [256, 256, 256, 512] \\
% 		\midrule
% 		DRL actor & [512, $\rm action\_dim$] \\
% 		DRL critic & [512, 512, 1] \\
% 		\bottomrule
% 	\end{tabular}
% \end{table*}


	
	
	
	% \end{thebibliography}
\end{document}